%% file: main.tex
\documentclass[11pt]{article}

\usepackage{acl}

\usepackage{times}
\usepackage{tcolorbox}

\usepackage[utf8]{inputenc} 
\usepackage[T1]{fontenc}   
\usepackage{hyperref}       
\usepackage{url}
\usepackage{multirow}       
\usepackage{booktabs}  
\usepackage{amssymb}
\usepackage{amsthm}

\usepackage{amsfonts}       
\usepackage{nicefrac}       
\usepackage{tabularx}
\usepackage{longtable}
\usepackage{supertabular}
\usepackage{makecell}
\usepackage{microtype}      
\usepackage{xcolor}         
\usepackage{tabularx}
\usepackage{graphicx}
\graphicspath{ {imgs/} }
\usepackage{svg}
\usepackage{enumitem}
\usepackage{subcaption}
\usepackage{amsthm, amsmath}
\usepackage[linesnumbered,ruled]{algorithm2e}
\usepackage{colortbl}

\definecolor{blue}{RGB}{17,220,247}
\definecolor{purple}{RGB}{163,115,250}
\definecolor{caribbeangreen}{rgb}{0.0, 0.8, 0.6}

\definecolor{GREEN}{RGB}{84,130,53}
\newcommand{\colorit}{\cellcolor{green!15}}
\newcommand{\coloritt}{\cellcolor{orange!15}}
\newcommand{\colorg}{\cellcolor{gray!15}}

\newcommand{\up}[1]{\tiny ($\textcolor{green}{\blacktriangle}#1\%)$}

\usepackage{wrapfig}

\usepackage{bbm}

\usepackage{algpseudocode}

 \DeclareMathOperator*{\argmax}{arg\,max}
 \DeclareMathOperator*{\argmin}{arg\,min}
 
\newcommand{\selfcon}{\textsc{sc}}

\newcommand{\static}{\textsc{explora}}
\newcommand{\staticshort}{\textsc{exp}}

\newcommand{\knn}{\textsc{knn}}
\newcommand{\mmr}{\textsc{mmr}}

\newcommand{\llamaseven}{\textsc{Llama-2-7B}}
\definecolor{GREEN}{RGB}{84,130,53}

\usepackage{bbm}

\usepackage{algpseudocode}

\newcommand{\mistral}{\textsc{Mistral-7B}}

\usepackage{pgfplots}
\pgfplotsset{compat=1.15}
\usepgfplotslibrary{
  statistics,
  colorbrewer,
  groupplots,
}
\usetikzlibrary{
  patterns,
  shapes.geometric,
  decorations.text,
  matrix,
  fit,
  backgrounds,
  positioning,
}
\tikzset{
  fignode/.style={
    outer sep=0.25em,
  }
}
\tikzset{
  framedfignode/.style={
    outer sep=0.25em,
    inner sep=0.5em,
    rounded corners,
    draw,
  }
}
\colorlet{plotColorNeutral}{gray}
\definecolor{plotColor1}{HTML}{f61a1c}
\definecolor{plotColor2}{HTML}{377eb8}
\definecolor{plotColor3}{HTML}{4daf4a}
\definecolor{plotColor4}{HTML}{984ea3}
\colorlet{plotColorNeutral*}{plotColorNeutral!40}
\colorlet{plotColor1*}{plotColor1!60}
\colorlet{plotColor2*}{plotColor2!60}
\colorlet{plotColor3*}{plotColor3!60}
\colorlet{plotColor4*}{plotColor4!60}
\pgfplotsset{
    colormap={greenred}{HTML=(4daf4a) HTML=(e41a1c)},
    colormap={redgreen}{HTML=(e41a1c) HTML=(4daf4a)}
}

\newcommand{\aqua}{\textsc{}{AquaRat}}
\newcommand{\tab}{\textsc{}{TabMWP}}
\newcommand{\strat}{\textsc{}{StrategyQA}}

\newcommand{\fin}{\textsc{}{FinQA}}
\newcommand{\gsm}{\textsc{}{GSM8K}}

\newcommand{\cX}{\mathcal{X}}

\newcommand{\cS}{\mathcal{S}}
\newcommand{\cU}{\mathcal{U}}
\newcommand{\ind}{\mathbbm{1}}
\newcommand{\bP}{\mathbb{P}}
\newcommand{\cV}{\mathcal{V}}
\newcommand{\cN}{\mathcal{N}}
\newcommand{\cL}{\mathcal{L}}
\newcommand{\cG}{\mathcal{G}}

\theoremstyle{definition}
\newtheorem{defin}{Definition}[section]

\title{EXPLORA: Efficient Exemplar Subset Selection for Complex Reasoning}

\author{Kiran Purohit \\ IIT Kharagpur \\ \texttt{kiran.purohit} \\\texttt{@kgpian.iitkgp.ac.in} \And Venktesh V \\ TU Delft \\\texttt{v.viswanathan-1} \\\texttt{@tudelft.nl}\And Raghuram Devalla \\ IIT Kharagpur \\ \texttt{devallaraghuram} \\ \texttt{@gmail.com} \AND Krishna Mohan Yerragorla \\ IIT Kharagpur \\\texttt{krishnamohanyerragorla}\\\texttt{@gmail.com } \And Sourangshu Bhattacharya \\ IIT Kharagpur \\ \texttt{sourangshu} \\ \texttt{@cse.iitkgp.ac.in } \And  Avishek Anand \\ TU Delft \\ \texttt{avishek.anand}\\ \texttt{@tudelft.nl}  }

\begin{document}
\maketitle
\begin{abstract}

\input{abstract.tex}

\end{abstract}

\input{intro-AA}
\input{related-work}
\input{03-framework}
\input{04-experiments}

\input{05-results}

\bibliography{references}
\bibliographystyle{acl_natbib}

\input{appendix}

\end{document}

%% file: abstract.tex
Answering reasoning-based complex questions over text and hybrid sources, including tables, is a challenging task. Recent advances in large language models (LLMs) have enabled in-context learning (ICL), allowing LLMs to acquire proficiency in a specific task using only a few demonstration samples (\textit{exemplars}).  A critical challenge in ICL is the selection of optimal exemplars, which can be either task-specific (static) or test-example-specific (dynamic). Static exemplars provide faster inference times and increased robustness across a distribution of test examples. In this paper, we propose an algorithm for static exemplar subset selection for complex reasoning tasks. We introduce \static{}, a novel exploration method designed to estimate the parameters of the scoring function, which evaluates exemplar subsets without incorporating confidence information. \static{} significantly reduces the number of LLM calls to $\sim$11\% of those required by state-of-the-art methods and achieves a substantial performance improvement of 12.24\%. We open-source our \textbf{code and data}\footnote{\url{https://github.com/kiranpurohit/EXPLORA}}. 




%% file: intro-AA.tex
\section{Introduction}
\label{sec:intro}

Answering complex questions that require multi-step reasoning \cite{chen2022finqa,tabmwp,aqua_rat,roy2022question:book,venktesh2024quantemp,v2024dexter} over structured and unstructured sources is an active research area with applications in finance, law, fact-checking and healthcare~\cite{wang2023survey,zhang2021explain}. 
Unlike fine-tuning task-specific models~\cite{chiang-chen-2019-semantically,aqua_rat,roy2016solving,cobbe2021training}, recent advances in Large Language Models (LLMs) have paved the way for new approaches that employ in-context learning (ICL) \cite{wei2023chainofthought} to solve complex reasoning problems. This approach focuses on choosing a small number of demonstration examples to be used as input prompts to LLMs.

An effective way to tackle complex reasoning problems is to use \textit{chain-of-thought} (COT) prompting, which adds \textit{hand-crafted natural language rationales} as stepwise solutions to the prompts, resulting in the triplet \texttt{(input, rationale, output)} \cite{wei2023chainofthought}. In this work, we refer to this triplet as an \textbf{exemplar}. A limitation of the COT-based method for reasoning tasks is the tedious and non-scalable manual effort required in selecting the rationales or \textit{exemplars}~\cite{lu-etal-2022-fantastically,zhao2021calibrate,chang2023data}.
To address these limitations, both static and dynamic approaches for automatic exemplar selection have been proposed~\cite{ye2023compositional,lu-etal-2022-fantastically,rubin-etal-2022-learning,complex_cot}.
Some approaches require annotated data and training of multiple models for exemplar selection \cite{lu2023dynamic,ye2023compositional}. 
Dynamic exemplar selection methods often involve additional computational costs because they select exemplars during query time, necessitating extensive query encoding and dynamic exploration of the search space. In contrast, static exemplar selection pre-selects a small subset of exemplars, which are used during LLM inference.
Prior exemplar selection methods do not capture interactions between the exemplars in the selected set \cite{li2023finding}. Additionally, current static selection approaches \cite{li2023finding} are characterized by a large number of LLM calls, which are computationally expensive.

In this work, we propose \static{}, a novel \textit{static exemplar subset selection} method that selects multiple low-loss exemplar subsets (overview in Figure \ref{fig:icl}). 
Our method is designed based on two hypotheses: (1) An effective exemplar selection algorithm for ICL should model the end-to-end ICL process, and (2) Prompt generators (refer Section \ref{sec:in-context}), which generate prompts using multiple exemplar subsets can be used to enhance the effectiveness of static ICL predictors for complex reasoning tasks.
Following these hypotheses, we model the problem of static exemplar selection for \textit{In-context Complex Reasoning} (ICCR) tasks as a novel top-$l$ \textit{exemplar subset-selection problem}. We use a linear model of similarity with validation examples for modeling the loss incurred by exemplar subsets.
We propose a novel sampling-based bandit algorithm for simultaneously estimating the parameters of the loss model and identifying the top-$l$ exemplar subsets, while incurring a low number of calls to the LLM (which corresponds to a low sample complexity of the bandit algorithm).
\static{} implicitly captures the interactions between the exemplars in the subset by scoring subsets.

We conduct extensive experiments across multiple reasoning-based QA tasks. 
Our results indicate that \static{} outperform both static and dynamic exemplar selection baselines by 12.24\% and 45.45\% respectively (Table \ref{tab:main_result}), while reducing the number of LLM calls to $\sim$ $11\%$ of the state-of-the-art~\cite{li2023finding} (Figure \ref{fig:llm_calls}).

\noindent\textbf{Contributions.} The contributions of our work are: 

    (1)  We propose a novel top-$l$ exemplar-subset selection approach, \static{}, for end-to-end in-context learning of complex reasoning tasks by approximating the loss for a given exemplar subset using a scoring function.  

    (2) We introduce a novel sampling-based bandit algorithm for efficiently learning the parameters of the scoring function and estimating the top-$l$ exemplar subsets. 

    (3) We demonstrate that the exemplars selected by \static{} on smaller LLMs can be well \textit{transferred} to larger LLMs (Table \ref{tab:llm_transfer}), reducing the cost incurred by larger LLMs for exemplar selection.

    (4) We show that exemplars selected by \static{} are more \textit{robust} compared to baselines in task performance (Table \ref{tab:robustness}).


%% file: related-work.tex
\section{Related Work}
\label{sec:rel-work}


While many existing techniques for complex reasoning tasks involve fine-tuning of specialized models \cite{chiang-chen-2019-semantically,amini-etal-2019-mathqa,chen-etal-2020-hybridqa}, these approaches require access to the model parameters. 
Recent developments in language models have introduced few-shot prompting approaches \cite{brown2020gpt3,wei2022emergent} through ICL~\cite{wei2023chainofthought,wang-etal-2023-plan,kojima2023large,chen2022program}
and COT in complex reasoning tasks~\cite{aqua_rat,cobbe2021training,few_shot, shin-etal-2020-autoprompt,v2023incontextabilitytransferquestion}.
However, a major drawback of these approaches is the need for manual selection of exemplars, which is tedious and non-scalable. 
ICL is also sensitive to the sample order \cite{lu-etal-2022-fantastically}, dataset, task, and models \cite{zhao2021calibrate} \cite{su2022selective}, making optimal exemplar selection essential for stable task performance.

\noindent\textbf{Exemplar-Selection for ICL:}
\label{sec:rel-exemplar-sel}
Several automated exemplar selection methods have been proposed to eliminate the need for manual selection. These include reinforcement learning-based approaches \cite{zhang-etal-2022-active,lu2023dynamic}, Determinantal Point Processes \cite{ye2023compositional}, Low Rank approximation (DQ-LoRe) \cite{xiong2023dq} and constrained optimization \cite{tonglet2023seer}, which are effective for reasoning tasks. Additionally, alternative learning-free methods for diverse exemplar selection, such as similarity-based \cite{rubin-etal-2022-learning}, complexity-based \cite{complex_cot} and MMR \cite{ye-etal-2023-complementary} have also been proposed. 
Existing dynamic exemplar selection methods incur additional computational costs during inference 
and mostly used for interpretability~\cite{zhang2021explain,anand2022explainable}.  
To address this, a small, representative set of exemplars can be selected for ICL. 
Unlike coreset selection methods \cite{guo2022deepcore}, which rely on gradient-based model updates, ICL performs the target tasks without any parameter updates.

To the best of our knowledge, there has been very little research in this area, with the closest work being LENS \cite{li2023finding}. 
LENS relies on LLM's output probabilities and cannot be extended to black-box LLMs. In this work, our approach for static exemplar selection is applicable to black-box and also other open language models. 

%% file: 03-framework.tex
\begin{figure*}[hbt!]
\centering

\includegraphics[width=0.76\textwidth]{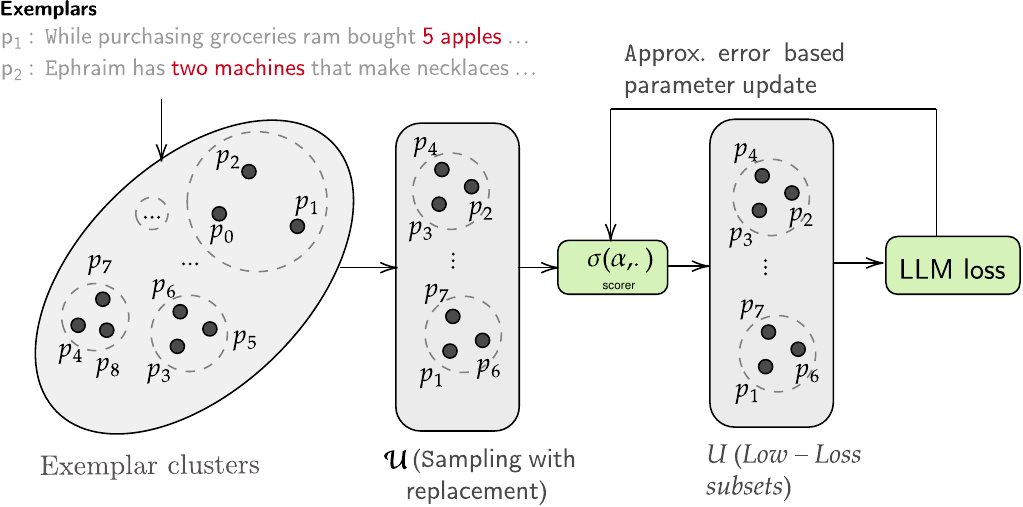}\\

\caption{\textbf{Overview of \static{}:} Initially, set $U$ is randomly selected from set $\cU$. In each iteration, parameters of the scoring function $\sigma(\alpha,.)$ are computed by minimizing a loss function. $\sigma$ guides the selection of the subset from $\cU\setminus U$ with the lowest loss, which is then used to update $U$. This iterative updating process ensures $U$ maintains low-loss subsets, leading to a more accurate estimation of $\alpha$ in subsequent iterations.}  
\label{fig:icl}
\end{figure*}

\section{\static{}: Model-based Exploration for Exemplar Subset Selection}
\label{sec:framework}
In black-box models, we cannot access the parameters of the LLMs or compute the gradient of the loss with respect to these parameters. Additionally, intermediate representations or generative probability scores from LLMs cannot be utilized for scoring exemplar subsets. To overcome these challenges, we introduce a novel approach to effectively select exemplar subsets without relying on the parameters of the LLMs.
Section \ref{sec:in-context} formally describes the components of ICL. 
Section \ref{sec:formulation} formulates the loss-model for exemplar subset selection.
Section \ref{sec:staticselection} motivates and describes the \static{} algorithm.

\subsection{ICL for Complex Reasoning}
\label{sec:in-context}

In-context learning (ICL) leverages LLMs to acquire proficiency in a specific task using only a few exemplars, without updating the model’s parameter. During this process, the LLM is provided with a prompt 
that includes \textit{input-rationale-output} triplet, referred to as \textit{exemplars}, which demonstrate the task to the LLM. 
Due to the financial and performance costs associated with large contexts, providing all $n$ training examples is impractical. Hence, exemplar selection methods curate a few exemplars that maximize overall accuracy.
Formally, let $\cX = \{x_i,z_i,y_i\}^{n}_{i=1} $ denote the set of all $n$ training examples (potential exemplars), and let $x_{test}$ be a test input. The goal is to predict the test output $y_{test}$. Let $S\subseteq\cX$ denote a subset of $k$ exemplars used for predicting $x_{test}$. 
The prompt $P$ is constructed as:\\
  $ P = [S, x_{test}] = [(x_{i_1},z_{i_1}, y_{i_1}), . . . \allowbreak, (x_{i_k},z_{i_1} \allowbreak , y_{i_k}),\allowbreak x_{test}] $, 
The end-to-end ICL process can be described as a composition of two steps: (1) response generator $f$, and (2) post-processing (decoding) $\delta$.
The response generator $f$ generates multiple responses, $r$ from the probability distribution $\bP_{LLM}$.
The post-processing step $\delta$ is applied to the LLM-generated response $f(P)$ to extract the task-specific output $\hat{y}_{test}$.
\begin{equation}
 \hat{y}_{test} = \delta(f([S, x_{test}]));\  f(P) = \cG\left(\bP_{LLM}(r|P)\right) \label{eq:ICL}
\end{equation}
Commonly used post-processing strategies include regular-expression matching ($\delta_{regex}$) and  self-consistency ($\delta_{SC}$) \cite{wang2022self}.

Reasoning problems \cite{2wikimultihopqa,aqua_rat} are especially challenging due to the complex relationship between the exemplars and the LLM's ability to perform multi-step reasoning tasks.
It has been demonstrated that providing rationales elucidating the reasoning steps improves LLM performance compared to state-of-the-art approaches \cite{wei2023chainofthought,complex_cot}.
Therefore, selecting exemplars with appropriate rationales tailored to the complex reasoning task is crucial \cite{xiong2023dq,tonglet2023seer}. However, generic exemplar selection approaches for ICL do not explicitly model this role of the rationales and interactions between exemplars.
End-to-end modeling of the entire ICL process (Eq \ref{eq:ICL}) is essential for capturing the complex role of the rationales.
In this work, we propose implicitly modeling the relationship between exemplars and explicitly modeling their relation to the LLM's performance by scoring subsets of exemplars (see section \ref{sec:formulation}).
This leads to formulation for the \textit{In-context Complex Reasoning} (ICCR) problem as an exemplar-subset selection problem.

A disadvantage of using a single subset of exemplars is that the prompt may not capture all diverse aspects of a given task.
This can be addressed using a prompt generator $\pi(S_1, ..., S_l, x_{test})$, which uses multiple subsets of exemplars, $S_1, ..., S_l$ and a test input $x_{test}$ to create a prompt $P$, improving overall performance and robustness compared to using a single exemplar subset.
For example, one can use a similarity-based prompt generator $\pi_{KNN}$, which selects semantically nearest neighbor subset, or a diversity-based prompt generator $\pi_{MMR}$ (the details are described in section \ref{sec:experiments}).
In this revised framework, the entire output generation process can be described as:
\begin{equation}
\hat{y}_{test} = \delta(f(P));\ P=\pi(S_1,...,S_l, x_{test}) 
\label{eq:predictive-model_static}
\end{equation}
In this setup, we are interested in finding a set $U=\{ S_1, ...., S_l \}$ of subsets such that the corresponding prompt $P$ generated by the prompt generator $\pi$ minimizes the total validation loss.
Let $\cV$ be the set of $m$ \textit{validation examples} $\{u_i, v_i\}^{m}_{i=1}$, where $u_i$ and $v_i$ represents the \textit{input} and the \textit{output} of the $i^{th}$ validation example respectively. We define the validation loss for a  prompt generator $\pi(U)$ with a set of exemplar subsets $U$ as:
\begin{align}
  L(\pi, U, \cV) = \frac{1}{m} \sum_{i=1}^m \ind(v_i \neq \delta(f(\pi(U, u_i)))) \label{eq:loss}
\end{align}
Hence, we define the problem of top-$l$ exemplar subset selection for ICCR as:
\begin{defin} [\textbf{ICCR}]
\textit{Let $\cV = \{u_i, v_i\}^{m}_{i=1}$ be a set of validation examples, and $U = \{S_1,...,S_l\}$ denote a set of subsets $S_i$ of size $k$. The problem of exemplar subset selection can be described as:
\begin{equation}
    U^* = \argmin_{U=\{S_1,...,S_l\}} L(\pi,U, \cV) \label{eq:exemplar-select}
\end{equation}
, $S_i\subseteq\cS$, $\cS$ is the set of all subsets $S$, $|S|=k$.}
\end{defin}

\color{black}


\subsection{Loss model for top-$l$ Exemplar Subset Selection}
\label{sec:formulation}

There are two main challenges in efficiently solving the exemplar subset selection problem (Eq \ref{eq:exemplar-select}):
(1) the set of all exemplar subsets $\cS$ can be very large leading to prohibitive time-complexity. For instance, $n=5000$ training examples, and a prompt size of $k=5$ examples lead to $C^5_{5000}$ (approximately $2.5 *10^{16}$) exemplar subsets. 
(2) a naive calculation of loss (Eq \ref{eq:loss}) for each exemplar subset $S$ involves $m$ calls ($m\sim 1000$) to the LLM, which can be expensive (both computationally and financially). We address the first issue in section \ref{sec:staticselection}. We propose to address the second issue by building a \textit{scoring function} ($\sigma$) for the subsets $S\in \cS$, which can be used to calculate the top-$l$ subsets without making calls to the LLM.

Intuitively, given a subset $S$, the scoring function $\sigma(S)$ should model the validation loss of an exemplar subset, $L(\pi,\{S\},\cV)$, since they are expected to generate identical rankings.
Hence, we propose that $\sigma$ should incorporate the relationship between the exemplar's question $x_i$ and the validation example's question $u_j\in \cV$.
In this work, we capture the relationship using a similarity score, $E_{ij} = \frac{\phi(x_i)^T\phi(u_j)}{\|\phi(x_i)\|\|\phi(u_j)\|} $, where $\phi(x)$ is the feature encoding of a smaller transformer model, e.g. BERT. 
We model score as linear function of the similarity features $E_{ij}$ of exemplars $x_i$ in the subset $S$:
\begin{align}
 \sigma(\Vec{\alpha}, S) = \frac{1}{m} \sum_{j=1}^m \sum_{i=1}^n \alpha_i \ind(x_i\in S) E_{ij}
\end{align}
Here, $\alpha_i$ denotes the $i$-th exemplar's contribution to the scoring function. We dynamically estimate $\alpha_i$ to fit the top-$l$ (lowest loss) exemplar subsets in $\cS$. Section \ref{sec:staticselection} describes the details of the algorithm.

Note that since $\alpha_i$'s can be negative, they can be learned to implicitly estimate both positive and negative correlations between training exemplars $x_i$ and $x_j$, according to signs of $\alpha_i$ and $\alpha_j$.
Also, note the above equation can be written as $ \sum_{i=1}^n \alpha_i \ind(x_i\in S) e_i$, where $e_i=\frac{1}{m} \sum_{j=1}^m E_{ij}$, which indicates that only aggregate effect of exemplars in validation set is modeled by $\sigma$.
Finally, training exemplars in ICL can be thought of as representing different ``distinguishable" solution ``concepts'' of tasks (see e.g. \cite{xie2021explanation}). 
For instance, consider the exemplars selected by \static{} for the AQUARAT dataset (Table \ref{tab:exemplar_qualitative_aquarat}) can be identified with concepts like \textit{Proportionality}, \textit{Series}, \textit{Kinematics}, and \textit{Interest}.
Since the final scoring function is trained to fit the top-$l$ exemplar subsets resulting in low loss, the resultant $\alpha_i$'s can be thought of as implicitly representing the important concepts needed for effectively predicting the correct label (hence resulting in low loss).
Unfortunately, since $\alpha_i$'s can be both positive and negative, their magnitudes are not directly interpretable as importance scores for the concepts.
Unlike existing works e.g. \cite{li2023finding, xu2024misconfidence}, we do not use probability scores for tokens provided by LLMs in our scoring function, as these scores are not directly related to the predictive task.

\begin{algorithm}[t]
  \SetAlgoLined

 \small
 \caption{\static{} }
 \label{algo:static_subsets}
 \small
 \textbf{Input:} $\cU \subseteq \cS$: \Comment{Initial exemplar subsets}\\
  \textbf{Initialize:}  \hspace{2mm}  $U_0 \leftarrow$ set of random $l$ subsets from $\cU$\\
   \hspace{2mm} $t \leftarrow 0$\\
   \hspace{2mm} $\Vec{\alpha} \leftarrow \cN(0,1)$ \Comment{Sampling from a gaussian } \\

\While{$t < T$} {
         Let $V_t \leftarrow \cU\setminus U_t$
        
            $ \Vec{\alpha_t} \leftarrow \min_{\Vec{\alpha}} \cL(\Vec{\alpha},U_t,V_t) $ \Comment{Eq. \ref{eq:alphaloss}}

         
          $S^*_t = \argmin_{S\in V_t}  \sigma(\Vec{\alpha_t},S)$ \Comment{Lowest loss subset}

         
          $\Tilde{S}_t = \argmax_{S\in U_t}   \sigma(\Vec{\alpha_t},S)$\Comment{Highest loss subset}
        
           
           \If{
           $\sigma(\Vec{\alpha_t},S^*_t)<\sigma(\Vec{\alpha_t},\Tilde{S}_t$)}
           {
           $U_{t} \leftarrow U_{t} \setminus \{\Tilde{S}_t\} $ \Comment{Remove $\Tilde{S}_t $}

           $U_{t+1} \leftarrow U_{t} \cup \{S^*_t\} $ \Comment{ add $S^*_t$}
           }
        
        
         $t \gets t+1$
   } 
   
 \textbf{Output:}
 $U_T$;Set of $l$ subsets from $\cU$ which have the lowest validation loss 
\end{algorithm}

\subsection{A Sampling-based bandit algorithm for Top-$l$ Exemplar Subset Selection}
\label{sec:staticselection}


\color{black}

Our exemplar-subset selection problem (Eq \ref{eq:exemplar-select}) in the context of the score function $\sigma$ encompasses two objectives: 
(1) learning the parameters of the loss model ($\sigma$) for the top-$l$ exemplar subsets, and
(2) calculating a set of low-loss exemplar subsets, $U$.
This problem can be posed as the top-$l$ arm selection problem in stochastic linear bandits \cite{kalyanakrishnan2012pac,chaudhuri2019pac}. 
In the current setting, the arms correspond to the exemplar subsets $S$, and feature vectors for the linear bandit are given by $E_{ij}$ where $x_i\in S$. The reward in our setting can be thought of as the negative of loss of a subset: $-L(\pi,\{S\},\cV)$.
\textit{LUCB} \cite{kalyanakrishnan2012pac,chaudhuri2019pac} and later generalized variants of GIFA \cite{reda2021top} are widely used for top-$l$ arm selection in stochastic linear bandits.  
\textit{LUCB} maintains two sets: (1) $l$ high-reward arms (called $U$ here), and (2) the other low-reward arms.
In each round, one arm is pulled from each of the sets, leading to a revised estimate of the rewards of all the arms.
However, these algorithms are impractical for our setting, since they require at least linear time in the number of arms, in each round. 


\input{table-results-SOTA}

Algorithm \ref{algo:static_subsets} describes \static{}, a novel sampling-based bandit algorithm, inspired by LUCB, for estimating $\alpha_i$ and identifying a set of $l$ low-loss (corresponding to high reward) exemplar subsets $U$.
\textit{For practicality}, we start with a manageable set $\cU\subseteq\cS$ of exemplar subsets after eliminating the obvious ones (see Section \ref{sec:experiments}).
We initialize $U_0$ as $l$ random subsets from $\cU$. 
$U_t$ in round $t$ denotes the set of $l$ subsets with the lowest loss. $V_t$ denotes the set of other subsets (note that we do not exhaustively enumerate $V_t$).
\static{} has two broad steps: (1) calculation of $\alpha_i$ based on losses computed using subsets from $U_t$ and $V_t$, and (2) updating $U_t$ based on the modified score function $\sigma$ (due to updation of $\alpha_i$).
$\alpha_i$ is updated by minimizing the following loss function:
\begin{align}
    \cL(\Vec{\alpha};U_t,V_t) &= \sum_{S\in U_t}  ( L(S, \cV) - \sigma(\Vec{\alpha},S))^2  \nonumber \\
    & +  \sum_{S' \sim V_t}  ( L(S', \cV) -  \sigma(\Vec{\alpha},S'))^2  
    \label{eq:alphaloss}
\end{align}
Here the first term denotes the approximation error of the loss model for the low-loss set $U_t$, and the second term denotes the loss on \textit{negative samples} from high-loss set $V_t$.
The negative samples facilitate exploration over the set $V_t$ by allowing the $\alpha_i$ values corresponding to unexplored and potentially low-loss subsets to be estimated correctly. 

The key motivation behind this formulation is to be able to frugally compute $\alpha_i$ while making minimal calls to the LLM. 
A naive computation of the first term requires $l*m$ calls to the LLM. However, since $U_{t+1}$ differs from $U_t$ by only one exemplar subset, a caching mechanism can implement this step in $m$ calls to the LLM, where $m$ is the validation set size.
The second term can be computed using $l'*m$ LLM calls, where $l'$ is the number of negative samples. 
$U_t$ is updated in lines 8 -- 12 in Algorithm \ref{algo:static_subsets}. $U_{t+1}$ differs from $U_t$ by only one exemplar subset. This leads to a smoother convergence of $\alpha_i$ over the iterations since the loss function $\cL$ depends mainly on $U_t$. Line 11 removes the exemplar subset $\Tilde{S}_t$ from $U_t$, which is the highest estimated loss subset in $U_t$, and line 12 adds $S^*_t$ to $U_t$, which is the exemplar subset with the lowest estimated loss in $V_t$. 
While a formal convergence guarantee for the proposed algorithm will be explored elsewhere, the updates is designed to decrease the total validation loss of $U_t$, provided that the estimation of loss $\sigma(\Vec{\alpha},S), S\in U_t$ becomes more accurate over the iterations. This can be achieved by reducing $l'$ over the rounds.
Also, note that the step in line 8 can be expensive to implement in many settings, due to the size of $V_t$. Here one can perform an approximate search that finds a ``good enough'' $S^*_t$ such that $\sigma(\Vec{\alpha},S^*_t) <  \sigma(\Vec{\alpha},\Tilde{S}_t)$.

\noindent

\color{gray}

\color{black}

%% file: table-results-SOTA.tex
\begin{table*}[t]
    \footnotesize
    \small

    \centering

     \resizebox{\textwidth}{!}{%
    \begin{tabular}{llllll}

    \toprule
     \textbf{Method}& \multicolumn{1}{l}{\textbf{GSM8K}}& \multicolumn{1}{l}{\textbf{AquaRat}} & \multicolumn{1}{l}{\textbf{TabMWP}} & \multicolumn{1}{l}{\textbf{FinQA}} 
     & \multicolumn{1}{l}{\textbf{StrategyQA}}  
     \\
     
    
    
    \midrule

    \coloritt  & \coloritt& \coloritt \textbf{GPT-3.5-turbo}& \coloritt & \coloritt& \coloritt\\

       \colorg \textbf{Dynamic} & \colorg & \colorg & \colorg & \colorg & \colorg \\
    KNN  \cite{rubin-etal-2022-learning}              &53.45    &51.96    &78.33   &51.52  & 81.83    \\
        KNN (S-BERT)  \cite{rubin-etal-2022-learning}              &53.07    &52.75    &77.95   &52.65  & 81.83    \\
    MMR  \cite{ye-etal-2023-complementary}            &54.36    &51.18    &77.32   &49.87    & 82.86  \\
    KNN+SC \cite{wang2022self}                        &80.21    &62.59    &83.08   &54.49 & 83.88     \\
    MMR+SC \cite{wang2022self}                        &78.01    &59.45    &81.36   &50.74  & 83.88    \\
    PromptPG \cite{lu2023dynamic}                     & \centering -       & -       &68.23   &53.56  & -    \\

    \colorg \textbf{Static} & \colorg& \colorg& \colorg& \colorg & \colorg \\
    Zero-Shot COT \cite{kojima2023large}              &67.02    &49.60    &57.10   &47.51 & 59.75         \\
    Manual Few-Shot COT \cite{wei2023chainofthought}  &73.46    &44.88    &71.22   &52.22 & 73.06     \\
    Random                                            &67.79    &49.80    &55.89   &53.70 &  81.02     \\
    PS+ \cite{wang-etal-2023-plan}                    &59.30    &46.00    & -      & -   & -      \\
    Auto-COT \cite{zhang2023automatic}                &57.10    &41.70    & -      & -  & 71.20       \\
    GraphCut \cite{iyer2013submodular}                &66.19    &47.24    &60.45   &52.31 & 80.00    \\ 
    FacilityLocation \cite{iyer2013submodular}        &68.61    &48.43    &67.66   &36.79 &  81.63    \\ 
    LENS \cite{li2023finding}                         &69.37    &48.82    &77.27  &54.75  & 79.79    \\ 
    LENS+SC \cite{li2023finding}                         &79.37    & 57.87   &  80.68 & 60.06   & 82.24  \\ 

     \colorg \textbf{Our Approach} & \colorg& \colorg& \colorg& \colorg & \colorg\\
     
    \textbf{\static{}}                 &\colorit 77.86\up{12.24} $\dagger$ &\colorit 53.54\up{9.67}$\dagger$ &\colorit 83.07\up{7.51} $\dagger$ &\colorit 59.46\up{8.60} $\dagger$ &\colorit 85.71\up{5.63} $\dagger$\\
    
    \textbf{\static{}+\selfcon{}}      &\textbf{86.35}\up{24.48} $\ddagger$&63.39\up{29.84} $\ddagger$  &85.52\up{10.68} $\ddagger$   &64.52\up{17.84}  $\ddagger$ & 87.14 \up{9.21}$\dagger$   \\
    
    \textbf{\static{}+\knn+\selfcon{}} &85.14 \up{22.73}$\ddagger$  & 62.20\up{27.41}$\ddagger$ & 86.29\up{12.39} $\ddagger$  &\textbf{65.12}\up{18.94}  $\ddagger$ & \textbf{88.37}\up{10.75}$\dagger$   \\
    
    \textbf{\static{}+\mmr+\selfcon{}} &86.13\up{24.16} $\ddagger$ &\textbf{63.78}\up{30.64} $\ddagger$&\textbf{86.96}\up{12.54}$\ddagger$    &64.60\up{17.99}   $\ddagger$    & 87.55\up{9.73}$\dagger$       \\

    \coloritt  & \coloritt& \coloritt \textbf{GPT-4o-mini}& \coloritt & \coloritt& \coloritt\\

    LENS \cite{li2023finding} &76.19 &64.56 &86.34 & 69.31 & 92.85    \\
   \textbf{\static{}} &93.63    &69.29    &90.12  &72.71  & 95.10    \\

     \bottomrule
    \end{tabular}
    } 
    \caption{Results across datasets (we use 5-shot for all methods). Percentage improvements are reported over LENS \cite{li2023finding}. $\dagger$ indicates statistical significance using t-test over LENS at 0.05 level and $\ddagger$  at 0.01 level.}
    \label{tab:main_result}
\end{table*}


%% file: 04-experiments.tex
\section{Experimental Setup}
 \label{sec:experiments}
We answer the following research questions.\\
\noindent\textbf{RQ I.} Can \static{} a static exemplar selection approach achieve competitive performance compared to the state-of-the-art?\\
\noindent\textbf{RQ II.} Can we transfer the exemplars selected with respect to smaller language models directly to Larger Language Models?\\
\noindent\textbf{RQ III.} Can we minimize the number of calls to the language models during exemplar selection?

\color{black}



\subsection{Experimental setting} 
\textbf{Datasets, Metrics}: We conduct extensive experiments over a range of complex reasoning datasets (\gsm{}, \aqua{}, \tab{}, \fin{} and \strat{}). 
We use official metrics of the datasets, i.e., \textit{exact match} (EM), cover-EM \cite{self_ask,cover_em}. More details about the datasets and prompts can be found in Appendix \ref{sec:datasets} and \ref{sec:prompts}.

\noindent\textbf{Hyperparameters}:
For all experiments, we set the temperature to $0.3$ to mitigate randomness, with frequency and presence penalty set to 0.8 and $0.6$ to avoid repetition. 
We set the max\_token\_length to 900 for generation. 
For efficiency reasons, we carry our experiments in a transfer setting, where we select exemplars using \static{} or other static exemplar selection methods for smaller models like Mistral-7b and Llama2-7b and then transfer them to a larger model like gpt-3.5-turbo to perform inference owing to its superior capabilities. We report performance on smaller LLMs in Appendix \ref{sec:small_llms}. 

\noindent\textbf{Subset selection hyperparameters}: For \static{}, we set $k$ as $5$, the desired number of clusters to be formed from the training set. We set the number of validation examples $\mathcal{V}$ to $20$. We construct a set $\cU$, of 40 subsets, each containing 5 exemplars, by randomly selecting one exemplar from each cluster with replacement. While we experimented with larger values for size of $\cU$ (100..etc.) we observe that at $\cU$=40, we achieve optimal performance on validation set. Initially, set $U$ consists of $10$ random subsets from set $\cU$, while $V$ comprises the remaining subsets not included in $U$. In each round, we randomly sample 5 subsets from $V$. We update $U$ by removing the worst subset and add the best subset from $V$ to it. This process repeats for $10$ iterations, with the stopping criterion of approximation error being unchanged between iterations, resulting in $U$ having 10 low-loss subsets. 

\noindent\textbf{\static{} Variants}:
 We posit that static and dynamic approaches can complement each other and apply dynamic methods like MMR and KNN over the $l$ subsets selected by \static{}, thereby reducing the search space. This makes \static{}+\knn{} and \static{}+\mmr{} a hybrid approach.

\subsection{Baselines}
\label{sec:baselines}








\textbf{Exemplar selection} : We compare with \textit{dynamic exemplar selection methods} like similarity (KNN) \cite{rubin-etal-2022-learning} and diversity (MMR) \cite{ye-etal-2023-complementary}. For KNN, we retrieve top 5 exemplars for each test example for a fair comparison with \static{}, and for MMR we observe $\lambda=0.5$ to be the optimal value. For KNN (S-BERT), we employ sentence transformer \textit{paraphrase-MiniLM-L6-v2}.
We also compare with the \textit{chain of thought methods} like Manual Few-Shot COT \cite{wei2023chainofthought}, Zero-Shot COT \cite{kojima2023large}, \textit{random} ,  \textit{coreset selection methods} (Facility Location and Graph Cut \cite{iyer2013submodular}) and task-specific approaches like  ``Plan and Solve" \cite{wang-etal-2023-plan} and Auto-COT \cite{zhang2023automatic}.

\textbf{LENS} \cite{li2023finding}: We compare with a closely related static exemplar selection method where the training data is filtered in two stages to extract informative examples.


%% file: 05-results.tex
\section{Results}
\label{sec:results}


\input{table-results-llms_transfer}

\subsection{Performance Comparison}

 To answer \textbf{RQ1}, we compare  \static{} and it's variants with state-of-the-art static and dynamic exemplar selection methods. We observe in Table \ref{tab:main_result} that \static{} outperforms the random baseline, which highlights the non-triviality of the proposed task of selecting task-level representative exemplars. We also observe that \static{} outperforms manual Few-Shot COT~\cite{wei2023chainofthought}. 

We also compare \static{} with LENS \cite{li2023finding} a static exemplar selection method for ICL. We observe that \static{} and its variants significantly outperform LENS. For instance, on GSM8K \static{} outperforms LENS by \textbf{12.24\%}. LENS scores each exemplar independently without considering any interactions between the exemplars, and also assumes access to LLM logits.
However, in \static{}, scores are assigned to subsets, allowing for the implicit capture of the interplay between the exemplars within each subset. This is particularly important for reasoning tasks, as the exemplars need to contain sufficient information for solving diverse reasoning based questions.  We perform a \textit{qualitative analysis} of exemplars chosen by LENS vs \static{} in \textbf{Appendix \ref{sec:exemplar_qualitative}}.
 
We also observe that existing coreset selection methods like Graph Cut and Facility Location perform worse than or are similar in performance to the random exemplar selection. This indicates the importance of designing methods specific to ICL for exemplar selection. 

 We also observe that \static{} outperforms dynamic exemplar selection methods like KNN, PromptPG and MMR.  
 A significant limitation, apart from additional inference time computational costs, is that dynamic exemplar selection methods do not consider interactions between the exemplars.

\input{table-results-robustness}

\subsection{ Transferability of exemplars from smaller LLMs to larger LLMs }
 To answer \textbf{RQ2}, we report the performance on test set in transfer setting across tasks using gpt-3.5-turbo with exemplars selected from Llama2-7b and Mistral-7b as shown in Table \ref{tab:llm_transfer}. In Table \ref{tab:main_result} we report the \static{} results from this transfer setup with exemplars selected from Mistral-7b.
 We observe that the exemplars selected by \static{} using smaller LLMs transfer well to larger LLMs, as indicated by their superior performance compared to baselines through evaluation in the transfer setting. This shows the strong transferability of our selected exemplars and the effectiveness of \static{}, which is robust across different LLMs. We attribute the transferability of the selected exemplars to design choice of \static{}, which remains agnostic to confidence scores from the LLMs.

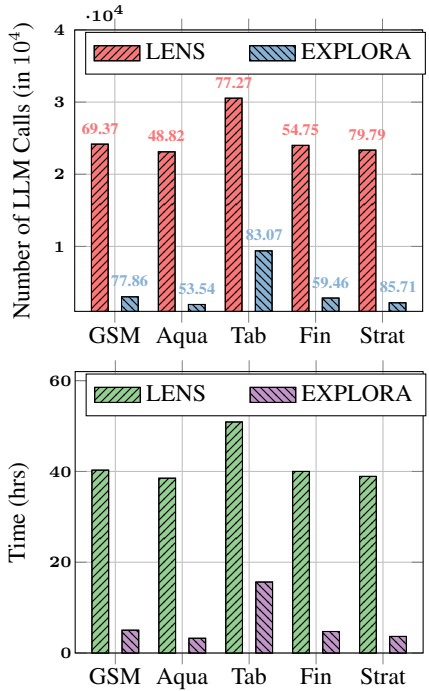
\begin{figure}[!t]
    \centering
    \begin{subfigure}{.8\linewidth}
    \centering
    \input{LLM_calls_plot.tex}
    \input{time_comparison.tex}
    \end{subfigure}
\caption{(Top) Frugal exemplar selection by \static{}: \textbf{LLM calls} LENS vs \static{} (y-axis) with corresponding EM scores indicated on top of bars. (Bottom) \textbf{Runtime} comparison LENS vs \static{}.}
\label{fig:llm_calls}
\end{figure}

\subsection{ Robustness of exemplars selected by \static{} compared to other approaches}

We compare the robustness of \static{} to other exemplar selection methods. We measure standard deviation of performance across different subsets of the evaluation set through 10-fold cross validation, as shown in Table \ref{tab:robustness}. We observe that the exemplars chosen by \static{} results in less variance across different subsets of the evaluation set when compared to other static exemplar selection methods. We also observe that in 3 out of 4 datasets, exemplars chosen by \static{} has less variance in task performance when compared to dynamic exemplar selection methods like KNN and MMR. Exemplars selected through dynamic approaches are not optimized for the task but rather on a per-test-example basis. Consequently, this leads to greater variance in final task performance. 
In \tab{}, we observe that the variance in results is low for all exemplar selection methods. Hence, \static{} helps select exemplars for the task which are more robust than other static methods or dynamic selection methods.







\subsection{Exemplar selection efficiency based on number of LLM evaluations}
\label{sec:llms}




To answer \textbf{RQ3}, we compare the number of calls made to the LLMs during the exemplar selection step. We compare the number of such calls and also running times for the LENS approach and our proposed approach \static{} as shown in Figure \ref{fig:llm_calls}. We observe that LENS has significantly more LLM calls than \static{} (about \textbf{10x}). This is because LENS relies on confidence estimates from the LLM for each training example and computes an informativeness score for all examples in the dataset, incurring expensive LLM calls for each example. Whereas, \static{} computes scores for whole exemplar subsets and employs a exploration based approach, resulting in LLM calls only for small number of subsets. In summary, \static{}  drastically reduces the number of LLM calls (to $\sim$ 11\% of calls made by LENS) and also reduces running time as shown in Figure \ref{fig:llm_calls} during exemplar selection step with significant performance gains.



\input{table-result-exhaustive-evaluation}
\subsection{Ablation Studies}
\label{sec:ablation}

We conduct an \textbf{exhaustive evaluation} to compare our exploration method and demonstrate its effectiveness, as shown in Table \ref{tab:exhaustive_evaluation}. Evaluating all possible subsets is infeasible, so we exhaustively evaluated a downsampled set of $40$ subsets ($\cU$ from \static{}) on the validation set. We then selected the subset with the minimum validation loss for subsequent inference on the test set. Our superior performance compared to the exhaustive evaluation is due to modeling the top-$l$ subsets in each round.

Additionally, we perform an ablation study where we fit the linear model once on the entire downsampled set of $40$ subsets (shown in Table \ref{tab:exhaustive_evaluation}). Then we select the subset with the minimum approximation error (loss) for subsequent inference on the test set. Note that in this ablation, the linear model is fit once for all subsets, whereas in \static{}, we model the top-$l$ subsets in each round.

\section{Conclusion}
In this work, we propose an efficient and robust task level exemplar subset selection method that identifies highly informative exemplar subsets. The proposed method saves resources by reducing the number of LLM calls, in contrast to the current state-of-the-art. We also observe that the exemplars obtained using smaller LLMs can be well transferred to larger LLMs.
\static{} outperforms existing static and dynamic exemplar selection methods. In future, we plan to further explore  hybrid exemplar selection and the impact of exemplars
for tasks involving complex reasoning.

\section{Acknowledgements}
We thank the reviewers and the meta-reviewer for their valuable and insightful feedback. We also thank the SERB grant "Online Subset Selection Algorithms for Data-centric Responsible AI and Efficient Computer Vision", sanction no. :  CRG/2023/004600  Dt. 18-02-202 for supporting this project.

\section{Contributions}
Kiran contributed to experimental design, idea conceptualization, execution, \textbf{core algorithm (\static{}) implementation}  and writing. Venktesh contributed to experimental design, idea conceptualization,  execution, writing. Raghuram and Krishna ran experiments on smaller LLMs. Sourangshu contributed to formulation and writing of Section 3. Sourangshu and Avishek mentored the project and contributed to idea conceptualization and writing.

\section{Limitations}
Our method deals with selecting top-$l$ exemplar subsets that are best suited to improve the overall performance through In-Context Learning (ICL). We identify certain limitations that could be addressed in future works.

 Our method is significantly efficient and computationally less resource intensive compared to state-of-the-art exemplar selection methods.  However, one of the limitations of our approach is that if the space of the subsets $\cU$ is large in some scenarios, then the computational time of the step (Algo \ref{algo:static_subsets}, line 8) that calculates the lowest loss subset from $V$ would increase. However, it would not increase the number of LLM calls and would still be computationally less resource intensive than the existing approaches. We would also need more efficient ways to sample negative examples (Eq \ref{eq:alphaloss}) with increase in size of $\cU$. While currently we propose a random sampling mechanism to sample negative examples from $V$, in future we plan to further analyze the impact of sampling negative examples for larger $\cU$ sizes. We defer this for future work, as it is beyond the scope of the current work.

 While \static{} converges as observed from our experiments, the current work does not provide an analysis or provable guarantees for convergence of the parameter $\alpha$. In future, we plan to provide an analysis for convergence of parameter $\alpha$.
 Also, it is unclear how our approach performs in the full retrieval and interactive retrieval settings~\cite{anand2023query}.
 In the future we intend to extend our approach to open-domain complex QA datasets~\cite{v2024dexter} 


\section{Ethical Considerations}
The intended use of the proposed approach is exemplar selection for reasoning problems that can be used to build QA systems for finance or education. Since our approach uses LLM for complex reasoning-based QA, the risks of hallucination \cite{hallucination} must be taken into consideration before deploying the approach. Since users may trust the hallucinated answers from the QA system, this may result in the spread of misinformation \cite{zhang2023ethical,albrecht2022despite}. We observe that \static{} is more robust across test instances compared to baselines due to the transfer of informative exemplars with rationales. Although hallucination is still a possibility when employing \static{} and the resulting QA systems are not infallible.

Additionally, we do not use any private information for the proposed approach. Though LLMs may have been pre-trained on sensitive information, our prompts do not elicit any sensitive information directly or indirectly.

%% file: table-results-llms_transfer.tex
\begin{table}[!t]
\footnotesize
    \centering
    \small
     \resizebox{\linewidth}{!}{%
    \begin{tabular}{lcccccc}
    \toprule
    
     \textbf{Method}& \multicolumn{1}{c}{\textbf{T}} & \multicolumn{1}{c}{\textbf{GSM}}& \multicolumn{1}{c}{\textbf{Aqua}} & \multicolumn{1}{c}{\textbf{Tab}} & \multicolumn{1}{c}{\textbf{Fin}}
     & \multicolumn{1}{c}{\textbf{Strat}} \\

    
    
    \midrule

    \textbf{\staticshort{}}                &  L    &79.07  &53.94  & 80.11     &  54.66  & 85.31    \\
                             &  M   &77.86  &53.54  & 83.07     &  59.46 & 85.71    \\
    \textbf{\staticshort{}+\selfcon{}}     &  L    &85.82  &63.78  & 86.76     &  61.16  & 85.10   \\
                             &  M   &86.35  &63.39  & 85.52     &  64.52  & 87.14    \\
    \textbf{\staticshort{}+\knn+\selfcon{}}&  L    &85.89  &64.17  & 85.74     &  63.64  & 86.53   \\
                             &  M   &85.14  &62.20  & 86.29     &  65.12 & 88.37    \\
    \textbf{\staticshort{}+\mmr+\selfcon{}}&  L    &86.20  &62.99  & 87.81     &  64.60  & 86.12   \\
                             &  M   &86.13  &63.78  & 86.96     &  64.60   & 87.55  \\

     \bottomrule
    \end{tabular}
    } 
    \caption{Results for transfer (T) of exemplars selected using \static{} (\staticshort{}) on smaller LLMs (Llama2-7b (L) and Mistral-7b (M)) to larger LLM (gpt-3.5-turbo).}
    \label{tab:llm_transfer}
\end{table}

%% file: table-results-robustness.tex
\begin{table}[!t]
    \footnotesize
    \centering
     \resizebox{\linewidth}{!}{%
    \begin{tabular}{lccccc}
    \toprule
    \small

\textbf{Datasets} & \textbf{GSM} & \textbf{Aqua} & \textbf{Tab} & \textbf{Fin} & \textbf{Strat}\\

    
    \midrule

    Zero-Shot COT    & $\pm$5.18   & $\pm$7.08     &  $\pm$1.84 & $\pm$4.50 & $\pm$4.19     \\
    Few-Shot COT  & $\pm$4.48  &  $\pm$12.03 & $\pm$1.66  &  $\pm$4.76 & $\pm$5.67   \\
    KNN   & $\pm$3.76   & $\pm$5.49  &  $\pm$1.27 &  $\pm$4.17  & $\pm$4.85 \\
    MMR    & $\pm$4.00  & $\pm$10.53  & $\pm$1.68  &  $\pm$6.10 &  $\pm$5.70  \\
    Graph Cut & $\pm$6.38& $\pm$8.18  & $\pm$2.03  &  $\pm$5.29 & $\pm$7.62 \\
    Facility Location & $\pm$4.23  & $\pm$6.71  & $\pm$1.74  &  $\pm$4.94 & $\pm$5.93 \\
    LENS & $\pm$5.04  & $\pm$6.67  & $\pm$1.59  &  $\pm$5.81 & $\pm$3.98 \\
    \textbf{\static{}} & $\pm$\textbf{3.39}  & $\pm$\textbf{4.93}  & $\pm$1.50  &  $\pm$\textbf{3.41} &$\pm$\textbf{3.95} \\
    
      \bottomrule
    \end{tabular}
    } 
    \caption{Comparison of robustness of \static{} to other approaches. We report standard deviation (lower is better) with scores from different splits of eval. set.}
    \label{tab:robustness}
\end{table}

%% file: LLM_calls_plot.tex
\begin{tikzpicture}
\edef\mylst{"69.37","48.82","77.27","54.75","79.79"}
\edef\explora{"77.86","53.54","83.07","59.46","85.71"}

    \begin{axis}[
            ybar=5pt,
            width=1\textwidth,
            bar width=0.25,
            every axis plot/.append style={fill},
            grid=major,
            xtick={1, 2, 3, 4, 5},
            xticklabels={GSM, Aqua, Tab, Fin, Strat},
            xlabel={},
            ylabel style = {font=\small},
        yticklabel style = {font=\boldmath \tiny,xshift=0.05ex},
        xticklabel style ={font=\small,yshift=0.5ex},
            ylabel={Number of LLM Calls (in $10^4$)},
            enlarge x limits=0.15,
            ymin=1000,
            ymax=40000,
            legend style ={font=\small,yshift=0.5ex},
            area legend,
            nodes near coords style={font=\tiny,align=center,text width=2em},
            legend entries={LENS, EXPLORA},
            legend cell align={left},
            legend pos=north west,
            legend columns=-1,
            legend style={/tikz/every even column/.append style={column sep=0.5cm}},
        ]
        \addplot+[
            ybar,
            plotColor1*,
            nodes near coords=\pgfmathsetmacro{\mystring}{{\mylst}[\coordindex]}\textbf{\mystring},
            nodes near coords align={vertical},
            draw=black,
            postaction={
                    pattern=north east lines
                },
        ] plot coordinates {
                (1,24176)
                (2,23111)
                (3,30543)
                (4,24004)
                (5,23347)
            };
        \addplot+[
            ybar,
            plotColor2*,
            draw=black,
            nodes near coords=\pgfmathsetmacro{\mystring}{{\explora}[\coordindex]}\textbf{\mystring},
    nodes near coords align={vertical},
            postaction={
                    pattern=north west lines
                },
        ] plot coordinates {
                (1,3019)
                (2,1954)
                (3,9386)
                (4,2847)
                (5,2190 )
            };

    \end{axis}

\end{tikzpicture}

%% file: time_comparison.tex
\begin{tikzpicture}
\edef\mylst{"","","","",""}
\edef\explora{"","","","",""}

    \begin{axis}[
            ybar=5pt,
            width=1\textwidth,
            bar width=0.25,
            every axis plot/.append style={fill},
            grid=major,
            xtick={1, 2, 3, 4, 5},
            xticklabels={GSM, Aqua, Tab, Fin, Strat},
            xlabel={},
            ylabel style = {font=\small},
        yticklabel style = {font=\boldmath \tiny,xshift=0.5ex},
        xticklabel style ={font=\small,yshift=0.5ex},
            ylabel={Time (hrs)},
            enlarge x limits=0.15,
            ymin=0,
            ymax=62,
            legend style ={font=\small,yshift=0.5ex},
            area legend,
            nodes near coords style={font=\tiny,align=center,text width=2em},
            legend entries={LENS, EXPLORA},
            legend cell align={left},
            legend pos=north west,
            legend columns=-1,
            legend style={/tikz/every even column/.append style={column sep=0.5cm}},
        ]
        \addplot+[
            ybar,
            plotColor3*,
            nodes near coords=\pgfmathsetmacro{\mystring}{{\mylst}[\coordindex]}\textbf{\mystring},
            nodes near coords align={vertical},
            draw=black,
            postaction={
                    pattern=north east lines
                },
        ] plot coordinates {
                (1,40.29)
                (2,38.51)
                (3,50.90)
                (4,40.00)
                (5,38.91)
            };
        \addplot+[
            ybar,
            plotColor4*,
            draw=black,
            nodes near coords=\pgfmathsetmacro{\mystring}{{\explora}[\coordindex]}\textbf{\mystring},
    nodes near coords align={vertical},
            postaction={
                    pattern=north west lines
                },
        ] plot coordinates {
                (1,5.03)
                (2,3.25)
                (3,15.64)
                (4,4.74)
                (5,3.65)
            };

    \end{axis}

\end{tikzpicture}

%% file: table-result-exhaustive-evaluation.tex
\begin{table}[!t]
    \footnotesize
    \centering
     \resizebox{\linewidth}{!}{%
    \begin{tabular}{lccccc}
    \toprule
    \small

\textbf{Datasets} & \textbf{GSM} & \textbf{Aqua} & \textbf{Tab} & \textbf{Fin} & \textbf{Strat}\\

    \midrule

    Exhaustive eval &76.72   &50.39   &  82.24 &  57.02 & 82.45\\

    \static{} (-exploration) & 75.89& 50.00& 75.16 & 50.30 & 80.40\\
    \ \hspace{4em} (Mistral) & \\
    
    \textbf{\static{} (Llama)} & \textbf{79.07}  & \textbf{53.94}  & 80.11  &  54.66 & \textbf{85.31} \\

    \textbf{\static{} (Mistral)} & \textbf{77.86}  & \textbf{53.54}  & \textbf{83.07}  &  \textbf{59.46} & \textbf{85.71} \\
    
      \bottomrule
    \end{tabular}
    } 
    \caption{Ablation studies: exhaustive evaluation, w/o exploration vs proposed exploration (\static{}).}
    \label{tab:exhaustive_evaluation}
\end{table}

%% file: appendix.tex
\newpage

\clearpage

\appendix

\input{table-datasets}

\section{Datasets Description}
\label{sec:datasets}

An overview of the dataset statistics and examples are shown in Table \ref{tab:datasets_overview}.

\textbf{FinQA}: Comprises financial questions over financial reports that require numerical reasoning with structured and unstructured evidence. Here, 23.42\% of the questions only require the information in the text to answer; 62.43\% of the questions
only require the information in the table to answer;
and 14.15\% need both the text and table to answer. Meanwhile, 46.30\% of the examples have
one sentence or one table row as the fact; 42.63\%
has two pieces of facts; and 11.07\% has more than
two pieces of facts. This dataset has 1147 questions in the evaluation set.

\textbf{\aqua{}}: This dataset comprises 100,000 algebraic word problems in the train set with dev and test set each comprising 254 problems. the problems are provided along with answers and natural language rationales providing the step-by-step solution to the problem. An examples problem is shown in Table \ref{tab:datasets_overview}.

\textbf{\tab{}}: It is a tabular-based math word problem-solving dataset with 38,431 questions. \tab{} is rich in diversity, where 74.7\% of the questions in TabMWP belong to free-text questions, while 25.3\% are multi-choice. We treat all questions as free-form type and do not provide any options to the LLM for consistent evaluation. We evaluate on the test set with 7686 problems.

\textbf{\gsm{}}: This dataset consists of linguistically diverse math problems that require multi-step reasoning. The dataset consists of 8.5K problems and we evaluate on the test set of 1319 questions.

\textbf{\strat{}}: To prove the generality of our approach for reasoning tasks, we evaluate on StrategyQA \cite{strategy_qa}, a dataset with implicit and commonsense reasoning questions. Since there is no public test set with ground truth answers, we perform stratified sampling done on 2290 full train set to split into 1800 train and 490 test.

\textbf{Metrics}: For TabMWP and StrategyQA we employ cover-EM \cite{cover_em,self_ask}, a relaxation of Exact Match metric which checks whether the ground truth answer is contained in the generated answer. This helps handle scenarios where LLM generates "4 hours" and the ground truth is "4". For other numerical reasoning datasets, we employ Exact match.

\section{Results using Alternate Open Source LLMs}
\label{sec:small_llms}
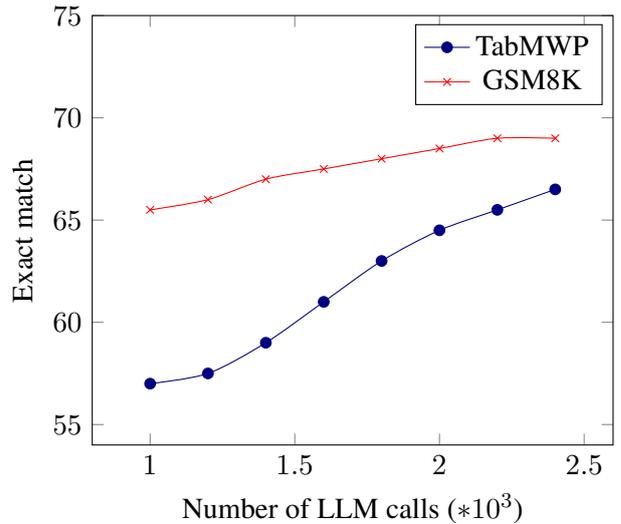
\begin{figure}
    \centering
        \input{line-plot}
    \caption{Plot showing number of LLM calls vs Exact Match for \tab{} and \gsm{}}
    \label{fig:line-plot}
\end{figure}

We also report the performance of the proposed exemplar selection approach \static{} on open-source models like Mistral-7b and LLama2-7b. The results are shown in Table \ref{tab:main_result_small_models}. We observe that the absolute  performance across baselines and \static{} is lower than when employing gpt-3.5-turbo  as backbone for the same exemplars. We primarily observe that this is due to the scale of the Language models as Mistral and LLAMA2 models have 7 billion parameters while gpt-3.5-turbo is of much larger scale and the emergent capabilities like In-Context Learning are proportional to scale of the language models \cite{wei2022emergent}.

However, we still observe that \static{} leads to reasonable performance gains over other static exemplar selection methods across the smaller open-source LLMs. We also observe that \static{} and its variants are competitive with dynamic exemplar selection methods.

Our main experiments are carried out in a  transfer setting where exemplar selection is done using small open source LLMs and transferred to larger LLMs. This is done for reducing the cost of LLM inference during exemplar selection, and also to leverage superior performance of LLMs with larger scale during inference. This setting is inspired from the work $\mu$P \cite{yang2022tensor} where the language model hyperparameters are tuned on a smaller LM and transferred to a larger language model.

\section{Prompts}
\label{sec:prompts}
We also demonstrate the instructions issued to the LLM for different tasks discussed in this work, along with some exemplars selected using \static{}. An example of prompt construction for \fin{} is shown in Figure \ref{prompt:finqa}. We also showcase example prompts for \aqua{} (Figure \ref{prompt:aqua}), \gsm{} (Figure \ref{prompt:gsm8k}), \tab{} (Figure \ref{prompt:tabmwp}) and \strat{} (Figure \ref{prompt:strategy}).

\section{Exemplar Qualitative Analysis}
\label{sec:exemplar_qualitative}
We provide a qualitative analysis of exemplars and compare the exemplars selected using \static{} with exemplars selected using LENS \cite{li2023finding}, the recent state-of-the-art approach. 

The final set of exemplars chosen by LENS vs \static{} for the \aqua{} dataset is shown in Table \ref{tab:exemplar_qualitative_aquarat}. We observe that Question 4 and Question 5 in the set of exemplars chosen by LENS are redundant in that they are very similar problems that require similar reasoning steps and are also similar thematically. Both the questions are centered on the theme of work and time and are phrased in a  similar manner. Hence, they do not add any additional information to solve diverse problems the LLM may encounter during inference. However, we observe that the exemplars chosen by \static{} are problems that require diverse reasoning capabilities and are also different thematically.

We also compare the exemplars chosen by \static{} with LENS for the \fin{} dataset ( Table \ref{tab:exemplar_qualitative_fin}) and make similar observations. We observe that the exemplars chosen by \static{} comprises diverse set of problems with diverse reasoning. We also observe that \static{} also contains exemplars that require composite numerical operations with multi-step reasoning rationales to arrive at the solutions, whereas LENS mostly has exemplars with single-step solutions.

The exemplars chosen by LENS compared to \static{} for TabMWP are shown in Table \ref{tab:exemplar_qualitative_tabmwp}. We observe that exemplar 1 and exemplar 3 chosen by LENS are redundant, as they represent the same reasoning concept of computing median for a list of numbers. However, we observe that \static{} selects diverse exemplars with each exemplar representing a different reasoning concept.

We also demonstrate the exemplars for GSM8K and StrategyQA in Table \ref{tab:exemplar_qualitative_gsm8k} and Table \ref{tab:exemplar_qualitative_strategyqa} respectively.


\section{Analysis of Accuracy (Exact Match) vs number of LLM calls}

We analyze the change in accuracy in proportion to number of calls to the LLM (\static{} Algo \ref{algo:static_subsets} iterations) as shown in Figure \ref{fig:line-plot}. We observe that the performance increases with number of LLM calls/iterations (Algorithm \ref{algo:static_subsets}) of \static{} algorithm. We also observe that for \gsm{} and \tab{}  \static{} converges and obtains optimal validation set performance quickly with less number of LLM calls, as observed in  Figure \ref{fig:line-plot}.

\input{table-results-small-models}

\input{prompt-aqua}

\input{prompt-finqa}
\input{prompt_gsm8k}
\input{prompt-tabmwp}
\input{prompt_strategy}

\input{exemplars-qualitative}

%% file: table-datasets.tex
\begin{table*}[b]
\resizebox{\textwidth}{!}{

\begin{tabular}{lccll}

\hline
\textbf{Dataset} & \textbf{\#Train} & \textbf{\#Test} & \textbf{Example Question} & \textbf{Description} \\ \hline

\colorg    & \colorg   & \colorg     & \colorg \texttt{Claire makes a 3 egg omelet every morning}      & multi-step  \colorg   \\
\colorg \textbf{GSM8K}~\cite{cobbe2021training}& \colorg 7473 & \colorg 1319 & \colorg  \texttt{for breakfast. How many dozens of eggs will}  & arithmetic word  \colorg    \\
\colorg & \colorg & \colorg & \colorg \texttt{she eat in 4 weeks?} & problems \colorg  \\

 &    &   &  \texttt{A trader sold an article at a profit of 20\%}      &   multi-step  \\
\textbf{AquaRat}~\cite{aqua_rat}& 97467 & 254 &  \texttt{for Rs.360. What is the cost price of the} & arithmetic word  \\
&  & & \texttt{article?} & problems  \\

\colorg   & \colorg    & \colorg   & \colorg  \texttt{Allie kept track of how many kilometers she}      & \colorg Table based   \\
\colorg \textbf{TabMWP}~\cite{tabmwp} & \colorg 23059 & \colorg 7686 & \colorg  \texttt{walked during the past 5 days. What is the} &\colorg numerical     \\
\colorg & \colorg & \colorg  & \colorg \texttt{range of the numbers?} &\colorg reasoning \\

 &    &   & \texttt{In 2010 and 2009, what was total fair value}      & Table and Text  \\
\textbf{FinQA}~\cite{chen2022finqa} & 6251 & 1147 & \texttt{in billions of assets segregated for benefit}      & based numerical  \\
& & & \texttt{of securities and futures brokerage customers?}      & reasoning
\\

\colorg   & \colorg    & \colorg   & \colorg  \texttt{Would the chef at Carmine's restaurant panic}      & \colorg multi-step   \\
\colorg \textbf{StrategyQA}~\cite{geva2021did} & \colorg 1800 & \colorg 490 & \colorg  \texttt{if there was no basil?} &\colorg reasoning     \\

\bottomrule
\end{tabular}
}
\caption{Overview of the Complex QA datasets used in this study.}
\label{tab:datasets_overview}
\end{table*}

%% file: line-plot.tex
\begin{tikzpicture}
\begin{axis}[
    xlabel=Number of LLM calls ($*10^3$),
    ylabel=Exact match,
    xmin=0.8, xmax=2.6,
    ymin=54, ymax=75]
    xtick={50,52,54,56...,75}
\addplot[smooth,mark=*,darkblue] plot coordinates {
    (1,57)
    (1.2,57.5)
    (1.4,59)
    (1.6,61)
    (1.8,63)
    (2.0,64.5)
    (2.2,65.5)
    (2.4,66.5)};
\addlegendentry{\tab{}}

\addplot[smooth,color=red,mark=x]
    plot coordinates {
    (1,65.5)
    (1.2,66)
    (1.4,67)
    (1.6,67.5)
    (1.8,68)
    (2.0,68.5)
    (2.2,69)
    (2.4,69)
    };
\addlegendentry{GSM8K}
\end{axis}
    \end{tikzpicture}

%% file: table-results-small-models.tex
\begin{table*}[hbt!]
    \centering
    \begin{tabular}{lcccc}
    \toprule
    
     \textbf{Method}& \multicolumn{1}{c}{\textbf{GSM8K}}& \multicolumn{1}{c}{\textbf{AquaRat}} & \multicolumn{1}{c}{\textbf{TabMWP}} & \multicolumn{1}{c}{\textbf{FinQA}} \\

    
    
    \midrule
    \colorg & \colorg & \colorg \textbf{Mistral-7B} & \colorg& \colorg \\\\
       \colorg \textbf{dynamic} & \colorg & \colorg & \colorg & \colorg   \\
    KNN  \cite{rubin-etal-2022-learning}             &37.98  &23.22    &61.74   & 9.06       \\
    MMR  \cite{ye-etal-2023-complementary}           &46.25  &18.11    &52.82   & 10.11      \\

    \colorg \textbf{static} & \colorg& \colorg& \colorg& \colorg\\
    Zero-Shot COT \cite{kojima2023large}             &7.43   &21.65    &43.34   &1.74        \\
    Manual Few-shot COT \cite{wei2023chainofthought} &30.48  &14.90    &46.94   &3.22        \\
    Random                                           &33.81  &22.04    &30.40   &5.52        \\
    GraphCut \cite{iyer2013submodular}               &47.00  &21.65    &59.46   &5.66        \\ 
    FacilityLocation \cite{iyer2013submodular}       &46.25  &14.17    &57.74   &4.96        \\ 
    LENS \cite{li2023finding}                        &46.39  &29.92    &57.69   &5.14        \\

    \colorg \textbf{Our Approach} & \colorg& \colorg& \colorg& \colorg \\
    \static{}                                        & 46.41  &30.07    &58.62   &5.57        \\
    \static{}+\selfcon{}                     &\textbf{61.41}  &\textbf{35.10}   &\textbf{61.96}   &7.41        \\

\midrule
    \colorg & \colorg& \colorg \textbf{Llama2-7B} & \colorg& \colorg \\\\
       \colorg \textbf{dynamic} & \colorg & \colorg & \colorg & \colorg  \\
    KNN  \cite{rubin-etal-2022-learning}             &23.43  &28.34    &54.83   & 10.37      \\
    MMR  \cite{ye-etal-2023-complementary}           &29.64  &21.65    &49.61   & 12.20      \\

    \colorg \textbf{static} & \colorg& \colorg& \colorg& \colorg \\
    Zero-Shot COT \cite{kojima2023large}             &6.14   &6.29     &40.31   &1.67        \\
    Manual Few-shot COT \cite{wei2023chainofthought} &21.15  &20.47    &43.23   &2.87        \\
    Random                                           &18.27  &21.60    &28.41   &4.62        \\
    GraphCut \cite{iyer2013submodular}               &27.29  &21.65    &47.38   &5.40        \\ 
    FacilityLocation \cite{iyer2013submodular}       &27.21  &21.65    &47.23   &5.05        \\ 
    LENS \cite{li2023finding}                        &28.05  &23.22    &48.29   &6.62        \\

    \colorg \textbf{Our Approach} & \colorg& \colorg& \colorg& \colorg \\
    \static{}                                        &28.67  &24.40    &49.96   &6.62        \\
    \static{}+\selfcon{}                             &\textbf{36.85}  &24.01    & \textbf{56.74}  &6.21        \\

     \bottomrule
    \end{tabular}
    \caption{Results across datasets on \mistral{} and \llamaseven{}.}
    \label{tab:main_result_small_models}
\end{table*}

%% file: prompt-aqua.tex
\begin{table*}[hbt!]

\begin{tcolorbox}[title= AQUA Prompt]
\small
\textbf{Instruction}:\texttt{You are a helpful, respectful and honest assistant helping to solve math word problems or tasks requiring reasoning or math. Follow given examples and solve the problems in step by step manner.}

\paragraph{\textbf{Exemplars}}:

[Question]: \textit{ The average age of three boys is 45 years and their ages are in proportion 3:5:7. What is the age in years of the youngest boy?}

[Options]: A) 9, B) 10, C) 11, D) 12, E) 13

[Explanation]: \textsf{$3x + 5x + 7x = 45$, \\ $x =3$, \\ $3x = 9$}

[Answer]: \textcolor{teal}{\textsf{The option is A}}
\\
\dots \\
\dots 
\paragraph{\textbf{Test Input}}: Question: {}
Options: {}

Explanation: [INS]
Answer: [INS]

\end{tcolorbox}
\captionof{figure}{Prompt for \aqua{}}
\label{prompt:aqua} 
\end{table*}









%% file: prompt-finqa.tex
\begin{table*}[hbt!]

\begin{tcolorbox}[title= FinQA Prompt]
\small
\textbf{Instruction}:\texttt{You are a helpful, respectful and honest assistant helping to solve math word problems or tasks requiring reasoning or math, using the information from given table and text.}

\paragraph{\textbf{Exemplars}}:







\textit{Read the following table, and then answer the question:}

[Table]: Year | 2016 | 2015 | 2014 |\\
share-based compensation expense | $30809$ | $21056$ | $29793$ | \\
income tax benefit | $9879$ | $6907$ | $7126$ |

[Question]: \textit{how much percent did the income tax benefit increase from 2014 to 2016?}

[Explanation]: \textsf{$x0=( 9879 - 7126$ ), \\ ans=( $x0 / 7126$ )}

[Answer]: \textcolor{teal}{\textsf{The answer is increased 38.6\%}}
\\
\dots \\
\dots 
\paragraph{\textbf{Test Input}}: Read the following table, and then answer the question:
Table: {}
Question: {}

Explanation: [INS]
Answer: [INS]

\end{tcolorbox}
\captionof{figure}{Prompt for \fin{}}
\label{prompt:finqa} 
\end{table*}

%% file: prompt_gsm8k.tex
\begin{table*}[hbt!]

\begin{tcolorbox}[title= GSM8K Prompt]
\small
\textbf{Instruction}:\texttt{You are a helpful, respectful and honest assistant helping to solve math word problems or tasks requiring reasoning or math. Follow given examples and solve the problems in step by step manner.}

\paragraph{\textbf{Exemplars}}:

[Question]: \textit{Samir just turned half the age Hania was $10$ years ago. If in five years Hania will be $45$ years old, what will Samir's age be five years from now?}

[Explanation]: \textsf{If in five years, Hania will be $45$ years old, currently she is $45-5 = 40$ years old.\\
Samir just turned half the age Hania was $10$ years ago, which means she is $30/2 = 15$ years old. \\
In five years, Samir will be $15+5 = 20$ years old.}

[Answer]: \textcolor{teal}{\textsf{20 years old}}
\\
\dots \\
\dots 
\paragraph{\textbf{Test Input}}: Question: {}

Explanation: [INS]
Answer: [INS]

\end{tcolorbox}
\captionof{figure}{Prompt for \gsm{}}
\label{prompt:gsm8k} 
\end{table*}

%% file: prompt-tabmwp.tex
\begin{table*}[hbt!]

\begin{tcolorbox}[title= TabMWP Prompt]
\small
\textbf{Instruction}:\texttt{You are a helpful, respectful and honest assistant helping to solve math word problems or tasks requiring reasoning or math, using the information from the given table. Solve the given problem step by step providing an explanation for your answer.}

\paragraph{\textbf{Exemplars}}:

[Table]: Table: Day | Number of tickets \\
Monday | $36$ \\
Tuesday | $43$ \\
Wednesday | $46$ \\
Thursday | $59$ \\
Friday | $37$ \\
Saturday | $46$ \\
Sunday | $51$

[Question]: \textit{The transportation company tracked the number of train tickets sold in the past 7 days. What is the range of the numbers?}

[Explanation]: \textsf{Read the numbers from the table. $36$, $43$, $46$, $59$, $37$, $46$, $51$ \\ First, find the greatest number. The greatest number is $59$. \\ Next, find the least number. The least number is $36$. \\ Subtract the least number from the greatest number: $59$ - $36$ = $23$}

[Answer]: \textcolor{teal}{\textsf{The range is 23}}
\\
\dots \\
\dots 
\paragraph{\textbf{Test Input}}: Table: {}
Question: {}

Explanation: [INS]
Answer: [INS]

\end{tcolorbox}
\captionof{figure}{Prompt for \tab{}}
\label{prompt:tabmwp} 
\end{table*}

%% file: prompt_strategy.tex
\begin{table*}[hbt!]

\begin{tcolorbox}[title= StrategyQA Prompt]
\small
\textbf{Instruction}:\texttt{You are a helpful, respectful and honest assistant helping to solve commonsense problems requiring reasoning. Follow the given examples that use the facts to answer a question by decomposing into sub-questions first and then predicting the final answer as "Yes" or "No" only.}

\paragraph{\textbf{Exemplars}}:

[Facts]: Snowden scored above $145$ on two separate IQ tests. The minimum accepted IQ score for MENSA on the Stanford–Binet is $132$, while for the Cattell it is $148$.

[Question]: \textit{Could Edward Snowden join MENSA?}

[Sub-question 1]: \textsf{What is the minimum accepted IQ score to be admitted to MENSA?\\}
[Sub-question 2]: \textsf{What is Edward Snowden's IQ?\\}
[Sub-question 3]: \textsf{Is \#2 greater than or equal to \#1?}

[Answer]: \textcolor{teal}{\textsf{Yes}}
\\
\dots \\
\dots 
\paragraph{\textbf{Test Input}}: Facts: {} Question: {}

Sub-question: [INS]
Answer: [INS]

\end{tcolorbox}
\captionof{figure}{Prompt for \strat{}}
\label{prompt:strategy} 
\end{table*}

%% file: exemplars-qualitative.tex
\begin{table*}
\begin{tabular}
{lp{.85\textwidth}}
\toprule
    \textbf{Method} & \textbf{Exemplars} \\
\midrule
\small

LENS  & \textbf{Question}: A cat chases a rat 6 hours after the rat runs. cat takes 4 hours to reach the rat. If the average speed of the cat is 90 kmph, what s the average speed of the rat?
\\ & \textbf{Options}: ['A)32kmph', 'B)26kmph', 'C)35kmph', 'D)36kmph', 'E)32kmph']
\\ & \textbf{Rationale:} Cat take 10 hours and rat take 4 hours...then Distance is 90*4.so speed of rat is (90*4)/10 = 36kmph 
\textbf{Answer: D}
 \\\cline{2-2}
 & \textbf{Question:}  A business executive and his client are charging their dinner tab on the executive's expense account.The \dots?
 \textbf{Options}: ['A)69.55\$', 'B)50.63\$', 'C)60.95\$', 'D)52.15\$', 'E)53.15']  \textbf{Rationale}:  let x is the cost of the food
1.07x is the gross bill after including sales tax
1.15* 1.07x=75 \textbf{Answer}: C \\\cline{2-2}
 & \textbf{Question:}John and David were each given X dollars in advance for each day they were expected to perform at a community festival. John eventually,\dots? 
 \\ &\textbf{Options}: 'A)11Y', 'B)15Y', 'C)13Y', 'D)10Y', 'E)5Y' \textbf{Rationale:} \dots  \textbf{Answer:} A\\\cline{2-2}
  & \textbf{Question:}A contractor undertakes to do a piece of work in 40 days. He engages 100 men at the beginning and 100 more after 35 days and completes the work in stipulated time. If he had not engaged the additional men, how many days behind schedule would it be finished?? \textbf{Options}: 'A)2', 'B)5', 'C)6', 'D)8', 'E)9'
  \textbf{Rationale:}  [(100$\times$ 35)+(200$\times$ 5)] men can finish the work in 1 day
therefore 4500 men can finish the work in 1 day. 100 men can finish it in  $\frac{4500}{100}$ = 45 days.
This is 5 days behind Schedule 
  \textbf{Answer:} A\\\cline{2-2}
    & \textbf{Question:} A can do a job in 9 days and B can do it in 27 days. A and B working together will finish twice the amount of work in ------- days? \\
 & \textbf{Options}: 'A)22 days', 'B)18 days', 'C)22 6/2 days', 'D)27 days', 'E)9 days'
  \textbf{Rationale:} 1/9 + 1/27= 3/27 = 1/9
9/1 = 9*2 =18 day
  \textbf{Answer:} B\\
  \midrule

\colorg \static{} & \colorg \textbf{Question}: The average age of three boys is 15 years and their ages are in proportion 3:5:7. What is the age in years of the youngest boy?
\textbf{Options}: ['A)9', 'B)10', 'C)11', 'D)12', 'E)13']
\textbf{Rationale: }3x + 5x + 7x = 45,
x =3,
3x = 9
\textbf{Answer: A}
 \\\cline{2-2}\colorg
 & \colorg\textbf{Question:}  Can you deduce the pattern and find the next number in the series? $6, 14, 26, 98$?
  \textbf{Options}: [''A)276', 'B)277', 'C)278', 'D)279', 'E)None of these'] \\\colorg
 & \colorg\textbf{Rationale}:  $6 = 1^1 + 2^1 + 3^1, 
14 = 1^2 + 2^2 + 3^2,
36 = 1^3 + 2^3 + 3^3,
98 = 1^4 + 2^4 + 3^4$
Thus the next number 
 \textbf{Answer}: A \\\cline{2-2}\colorg
 & \colorg\textbf{Question:}In covering a distance of 42 km, A takes 2 hours more than B. If A doubles his speed, then he would take 1 hour less than B. A's speed is:? 
\\ \colorg & \colorg \textbf{Options}: 'A)5 km/h', 'B)7 km/h', 'C)10 km/h', 'D)15 km/h', 'E)25 km/h' \textbf{Rationale:} Let A's speed be X km/hr.
Then, 42/x - 42/2x = 3
6x = 42
x = 7 km/hr \textbf{Answer:} B\\\cline{2-2}\colorg
  &\colorg \textbf{Question:}Find  the number which when multiplied by 15 is increased by 196. \\\colorg
 &\colorg \textbf{Options}: 'A)14', 'B)20', 'C)26', 'D)28', 'E)30'
  \textbf{Rationale:}  Solution
Let the number be x .
Then, 15x - x = 196
‹=›14x = 196
x ‹=› 14 
  \textbf{Answer:} A\\\cline{2-2}\colorg
    & \colorg\textbf{Question:} A certain sum of money at simple interest amounted Rs.980 in 3 years at 5\% per annum, find the sum? \textbf{Options}: 'A)867', 'B)855', 'C)299', 'D)852', 'E)903'
  \textbf{Rationale:} 980 = P [1 + (3*5)/100]
P = 852 
  \textbf{Answer:} D\\
\midrule
\midrule
\end{tabular}
\caption{Qualitative analysis of exemplars for \textbf{\aqua{}} dataset selected by LENS vs \static{}. Rationale is not completely shown for some questions to conserve space. However, in our experiments all exemplars include rationales.}
\label{tab:exemplar_qualitative_aquarat}
\end{table*}

\begin{table*}
\begin{tabular}{lp{.85\textwidth}}
\toprule
    \textbf{Method} & \textbf{Exemplars} \\
\midrule
\small

LENS  & \textbf{Table}: | increase ( decrease ) |
average yield | 2.75\% ( 2.75 \% ) |
volume | 0.0 to 0.25 |
energy services | 2013 |
fuel recovery fees | 0.25 |
recycling processing and commodity sales | 0.25 to 0.5 |
acquisitions / divestitures net | 1.0 |
total change | 4.25 to 4.75\% ( 4.75 \% ) |
\textbf{Question}: what is the ratio of the acquisitions / divestitures net to the fuel recovery fees as part of the expected 2019 revenue to increase
\\ & \textbf{Rationale:} ans=( 1.0 / 0.25 )
\textbf{Answer: }  The answer is 4
 \\\cline{2-2}
 & \textbf{Table:}  ( in millions ) | 2009 | 2008 | 2007 |
sales and transfers of oil and gas produced net of production andadministrative costs |  -4876 ( 4876 ) |  -6863 ( 6863 ) |  -4613 ( 4613 ) |
\dots
 \\ & \textbf{Question}: were total revisions of estimates greater than accretion of discounts?
\\
 & \textbf{Rationale}: \dots   \textbf{Answer}: The answer is yes  \\\cline{2-2}
 & \textbf{Table:} | 2007 | 2008 | change |
capital gain distributions received | 22.1 |  5.6 | -16.5 ( 16.5 ) |
other than temporary impairments recognized | -.3 ( .3 ) | -91.3 ( 91.3 ) | -91.0 ( 91.0 ) |
net gains ( losses ) realized onfund dispositions | 5.5 | -4.5 ( 4.5 ) | -10.0 ( 10.0 ) |
net gain ( loss ) \dots
\textbf{Question}: what percentage of tangible book value is made up of cash and cash equivalents and mutual fund investment holdings at december 31 , 2009?
\textbf{Rationale:} ( 1.4 / 2.2 )  \textbf{Answer:}  The answer is 64\%\\\cline{2-2}
  & \textbf{Table:} in millions | 2009 | 2008 | 2007 |
sales | 5680 | 6810 | 6530 |
operating profit | 1091 | 474 | 839 |
 \textbf{Question}: north american printing papers net sales where what percent of total printing paper sales in 2009?
   \textbf{Rationale:} x0=( 2.8 * 1000 ), ans=( x0 * 5680 )
  \textbf{Answer:} The answer is 49\%\\\cline{2-2}
    & \textbf{Table:}  in millions | december 312015 | december 312014 |
total consumer lending | 1917 | 2041 |
total commercial lending | 434 | 542 |
total tdrs | 2351 | 2583 |
nonperforming | 1119 | \dots
 \textbf{Question}: what was the change in specific reserves in all between december 31 , 2015 and december 31 , 2014 in billions?
  \textbf{Rationale:} ( .3 - .4 )
  \textbf{Answer:}  The answer is -0.1\\
  \midrule

\colorg \static{} & \colorg \textbf{Table}: | 2008 | 2007 |
balance at beginning of year | 23.2 | 56.4 |
additions due to acquisition of allied | 582.9 | 2014 |
additions based on tax positions related to current year | 10.6 | 16.3 |
reductions for tax positions related to the current year | -5.1 ( 5.1 ) | -17.2 ( 17.2 ) |
\dots
\textbf{Question}: in 2008 what was the change in the gross unrecognized tax benefits in millions
\textbf{Rationale:}  (611.9 - 23.2 )
\textbf{Answer: } The answer is 588.7
 \\\cline{2-2}\colorg
 & \colorg\textbf{Table:} december 31 2004 | 1054 |
december 31 2005 | 1216 |
december 31 2006 | 1219 |
december 31 2007 | 2566 |
  \textbf{Question}: what was devon's average translation adjustments included in accumulated other comprehensive income ( in millions ) from 2004 through 2007?
\textbf{Rationale}: x0=( 1054 + 1216 ), x1=( x0 + 1219 ), x2=( x1 + 2566 ), ans=( x2 + 4 )
 \textbf{Answer}: The answer is 1513.75 \\\cline{2-2}\colorg
 & \colorg\textbf{Table:} | 2016 | 2015 ( in thousands ) | 2014 |
share-based compensation expense | 30809 | 21056 | 29793 |
income tax benefit | 9879 | 6907 | 7126 |
\\ \colorg & \colorg \textbf{Question}:  how much percent did the income tax benefit increase from 2014 to 2016?
  \\\colorg  & \colorg  \textbf{Rationale:}  x0=( 9879 - 7126 ), ans=( x0 - 7126 )
 \textbf{Answer:} The answer is increased 38.6\%\\\cline{2-2}\colorg
  &\colorg \textbf{Table:}  in billions | 2018 |
january 1 |  33.3 |
issuances | 4.5 |
calls and maturities | -6.8 ( 6.8 ) |
other | -.1 ( .1 ) |
december 31 | 30.9 |
\\\colorg
 &\colorg \textbf{Question}: assuming all matured securities were pledged as collateral , how much should we assume came from the calls?
  \textbf{Rationale:}  ans=( 6.8 - 4.9 ) 
  \textbf{Answer:} The answer is 1.9 \\\cline{2-2}\colorg
    & \colorg\textbf{Table:}  in millions of dollars | u.s . | outside u.s . | december 31 2008 | december 31 2007 |
commercial and similar letters of credit | 2187 | 6028 | 8215 | 9175 |
 \dots \textbf{Question}: what percentage of citigroup 2019s total other commitments as of december 31 , 2008 are outside the u.s.?
  \textbf{Rationale:}  ans=( 236931 / 1349500 )
  \textbf{Answer:} The answer is 18\%\\
\midrule
\midrule
\end{tabular}
\caption{Qualitative analysis of exemplars for \textbf{\fin{}} dataset selected by LENS vs \static{}. Rationale is not completely shown for some questions to conserve space. However, in our experiments all exemplars include rationales.}
\label{tab:exemplar_qualitative_fin}
\end{table*}

\begin{table*}
\begin{tabular}{lp{.85\textwidth}}
\toprule
    \textbf{Method} & \textbf{Exemplars} \\
\midrule
\small

LENS  & \textbf{Question}: Michael wants to dig a hole 400 feet less deep than twice the depth of the hole that his father dug. The father dug a hole at a rate of 4 feet per hour. If the father took 400 hours to dig his hole \dots ?
\textbf{Rationale:} Since the father dug a hole with a rate of 4 feet per hour, if the father took 400 hours digging the hole, he dug a hole 4*400 = 1600 feet deep. \dots Michael will have to work for 2800/4 = 700 hours.
\textbf{Answer:} 700
 \\\cline{2-2}
 & \textbf{Question:} When Erick went to the market to sell his fruits, he realized that the price of lemons had risen by $4$ for each lemon. The price of grapes had also increased by half the price that \dots?
 \textbf{Rationale}:  The new price for each lemon after increasing by $4$ is $8+4 =12$ For the 80 lemons, \dots Erick collected $140*9=1260$ From the sale of all of his fruits, Erick received $1260+960 =2220$.
  \textbf{Answer}: 2220 \\\cline{2-2}
 & \textbf{Question:}James decides to build a tin house by collecting 500 tins in a week. On the first day, he collects 50 tins. On the second day, he manages to collect 3 times that number. \dots? 
\textbf{Rationale:} On the second day, he collected 3 times the number of tins he collected on the first day, which is $3*50 =150$ tins. \dots he'll need to collect $200/4=50$ tins per day to reach his goal. \textbf{Answer:} 50\\\cline{2-2}
  & \textbf{Question:} Darrel is an experienced tracker. He can tell a lot about an animal by the footprints it leaves behind. Based on the impressions, he could tell the animal was traveling east at 15 miles per hour \dots ?  \textbf{Rationale:} If we let x be the amount of time, in hours, it will take for Darrel to catch up to the coyote, \dots If we subtract 1 x from each side, we get x=1, the amount of time in hours.
 \textbf{Answer:} 1\\\cline{2-2}
    & \textbf{Question:} Martha needs to paint all four walls in her 12 foot by 16 foot kitchen, which has 10 foot high ceilings \dots If Martha can paint 40 square feet per hour, how many hours will it take her to paint kitchen? 
\textbf{Rationale:} There are two walls that are 12' by 10' and two walls that are 16' by 10' \dots how many hours she needs to finish: 1680 sq ft / 40 sq ft/hour = 42 hours 
  \textbf{Answer:} 42\\
  \midrule

\colorg \static{} & \colorg \textbf{Question}: The difference between the number of boys and girls in a tree planting event is 400. If there are 600 boys at the event, and the number of girls is more than the number of boys \dots ?
\textbf{Rationale:}  If there are 600 boys at the event, and the difference between boys and girls is 400, there are 600+400=1000 girls.
\dots 60\% of the total number of boys and girls at the event is 60/100*1600=960
\textbf{Answer:} 960
 \\\cline{2-2}\colorg
 & \colorg\textbf{Question:}  Casey is trying to decide which employee she wants to hire. One employee works for $\$20$ an hour. The other employee works for $\$22$ an hour, but Casey would also get a $\$6$/hour subsidy \dots ?
\textbf{Rationale}:  First find the weekly cost of the first employee: \$20/hour * 40 hours/week = 800/week \dots subtract the smaller weekly cost from the bigger weekly cost: \$800/week - \$640/week = 160/week.
 \textbf{Answer}: 160 \\\cline{2-2}\colorg
 & \colorg\textbf{Question:}Cara has 60 marbles in a bag. 20 of them are yellow, half as many are green, \dots If Cara picks a marble at random, what are the odds it's blue (expressed as a percentage)?
\\ \colorg & \colorg \textbf{Rationale:} First find the number of green marbles: 20 yellow marbles/2=10 green marbles. \dots find the chance of getting a blue marble: 15/60 marbles*100=25\%.
\textbf{Answer:} 25\\\cline{2-2}\colorg
&\colorg \textbf{Question:} Samir just turned half the age Hania was 10 years ago. If in five years Hania will be 45 years old, what will Samir's age be five years from now?
\textbf{Rationale:} If in five years, Hania will be 45 years old, currently she is 45-5=40 years old. Ten years ago, Hania was 40-10=30 years old. \dots In five years, Samir will be 15+5=20 years old. 
\textbf{Answer:} 20\\\cline{2-2}\colorg
& \colorg\textbf{Question:} If the normal hours of operation of Jean's business are 4 pm to 10p every day Monday through Friday, and from 6 pm to 10 pm on weekends, how many hours is the business open in a week?
\textbf{Rationale:} First, we find the number of hours per weekday by subtracting the smaller number from the larger one \dots We add these two totals together to find that 30+8=38 hours per week.
\textbf{Answer:} 38\\
\midrule
\midrule
\end{tabular}
\caption{Qualitative analysis of exemplars for \textbf{\gsm{}} dataset selected by LENS vs \static{}. Rationale is not completely shown for some questions to conserve space. However, in our experiments all exemplars include rationales.}
\label{tab:exemplar_qualitative_gsm8k}
\end{table*}

\begin{table*}
\begin{tabular}{lp{.85\textwidth}}
\toprule
    \textbf{Method} & \textbf{Exemplars} \\
\midrule
\small

LENS  & \textbf{Table}: | Name | Age (years)
| Jessica | 2
| Dalton | 7
| Kelsey | 5
| Lamar | 8
| Alexis | 2
 \textbf{Question}: A girl compared the ages of her cousins. What is the median of the numbers?
\textbf{Rationale:} Read the numbers: 2, 7, 5, 8, 2.
Arrange the numbers in ascending: 2, 2, 5, 7, 8.
Find the number in middle. The number in middle is 5.
The median is 5.
\textbf{Answer: } 5
 \\\cline{2-2}
 & \textbf{Table:} | City | Number of houses sold 
| Melville | 878
| New Hamburg | 871
| Charles Falls | 881
| Pennytown | 817
\textbf{Question}: A real estate agent looked into how many houses were sold in different cities. Where were the fewest houses sold? \textbf{Rationale}: Find the least number in the table. \dots Pennytown corresponds to 817.
   \textbf{Answer}: 817  \\\cline{2-2}
 & \textbf{Table:} | Day | Number of new customers
| Saturday | 2
| Sunday | 2
| Monday | 9
| Tuesday | 4
| Wednesday | 10
| Thursday | 3
| Friday | 6
\textbf{Question}: A cable company paid attention to how many new customers it had each day. What is the median of the numbers?
\textbf{Rationale:} Find the number in middle. The number in middle is 4.
The median is 4.
  \textbf{Answer:} 4 \\\cline{2-2}
  & \textbf{Table:} | Day | Number of cups
| Friday | 8
| Saturday | 4
| Sunday | 10
| Monday | 6
| Tuesday | 6
| Wednesday | 1
| Thursday | 0
 \textbf{Question}: Nancy wrote down how many cups of lemonade she sold in the past 7 days. What is the range of the numbers?
\textbf{Rationale:} Subtract the least number from the greatest: 10-0=10. The range is 10.
\textbf{Answer:} 10\\\cline{2-2}
    & \textbf{Table:}  | Price | Quantity demanded | Quantity supplied
|\$700 | 9,800 | 22,600
|\$740 | 8,000 | 22,800
|\$780 | 6,200 | 23,000
|\$820 | 4,400 | 23,200
|\$860 | 2,600 | 23,400
\textbf{Question}: At a price of \$860, is there a shortage or a surplus?
 \textbf{Rationale:} At price of \$860, quantity demanded is less than quantity supplied. \dots So, there is a surplus.
 \textbf{Answer:}  surplus\\
 \midrule

\colorg \static{} & \colorg \textbf{Table}: | Day | Number of tickets
| Monday | 36
| Tuesday | 43
| Wednesday | 46
| Thursday | 59
| Friday | 37
| Saturday | 46
| Sunday | 51
\textbf{Question}: The transportation company tracked the number of tickets sold in the past 7 days. What is the range of the numbers?
\textbf{Rationale:} Find greatest number. \dots Subtract least from the greatest number: 59 - 36 = 23.
The range is 23.
\textbf{Answer: } 23
 \\\cline{2-2}\colorg
 & \colorg\textbf{Table:} | Stem | Leaf 
| 4 | 2, 7, 9, 9, 9
| 5 | 1
| 6 | 9
| 7 | 2, 2, 3, 3, 5
| 8 | 
| 9 | 0
\textbf{Question}:  A pottery factory kept track of the number of broken plates per shipment last week. How many shipments had exactly 73 broken plates?
\textbf{Rationale}:  For the number 73, the stem is 7, and the leaf is 3. Find the row where the stem is 7. In that row, count all the leaves equal to 3.
\dots 2 shipments had exactly 73 broken plates.
\textbf{Answer}: 2 \\\cline{2-2}\colorg
 & \colorg\textbf{Table:} | purple and red clay bead | \$0.02
| small pink bead | \$0.04
| pearl bead | \$0.07
| round silver bead | \$0.01
| brown cat's eye bead | \$0.08
| orange glass bead | \$0.07
\textbf{Question}:  Kylie has \$0.05. Does she have enough to buy a small pink and purple and red clay bead?
\textbf{Rationale:}  Add the price of a small pink and purple and red clay bead: \$0.04 + \$0.02 = \$0.06.
\$0.06 is more than \$0.05. \textbf{Answer:} Kylie does not have enough money.  \\\cline{2-2}\colorg
  &\colorg \textbf{Table:}  | Price | Quantity demanded | Quantity supplied
| \$165 | 17,900 | 6,400
| \$345 | 15,100 | 8,900
| \$525 | 12,300 | 11,400
| \$705 | 9,500 | 13,900
| \$885 | 6,700 | 16,400
\textbf{Question}: Look at the table. Then answer the question. At a price of \$885, is there a shortage or a surplus?
\textbf{Rationale:}  At the price of \$885, the quantity demanded is less than the quantity supplied. \dots So, there is a surplus.
\textbf{Answer:} surplus \\\cline{2-2}\colorg
& \colorg\textbf{Table:} | Chickenville | 3:00 A.M. | 12:00 P.M. | 3:30 P.M.
| Floral Gardens | 3:45 A.M. | 12:45 P.M. | 4:15 P.M.
| Pleasant River Campground | 4:45 A.M. | 1:45 P.M. | 5:15 P.M.
| Happy Cow Farm | 5:15 A.M. | 2:15 P.M. | 5:45 P.M.
| Rocky Ravine Town | 5:45 A.M. | 2:45 P.M. | 6:15 P.M.
\textbf{Question}: Look at the following schedule. Doug just missed the 3.00 A.M. train at Chickenville. How long does he have to wait until the next train?
\textbf{Rationale:}  Find 3:00 A. M. in the row for Chickenville. Look for the next train in that row.
The next train is at 12:00 P. M. The elapsed time is 9 hours.
\textbf{Answer:} 9\\
\midrule
\midrule
\end{tabular}
\caption{Qualitative analysis of exemplars for \textbf{\tab{}} dataset selected by LENS vs \static{}. Rationale is not completely shown for some questions to conserve space. However, in our experiments all exemplars include rationales.}
\label{tab:exemplar_qualitative_tabmwp}
\end{table*}

\begin{table*}
\begin{tabular}{lp{.85\textwidth}}
\toprule
    \textbf{Method} & \textbf{Exemplars} \\
\midrule
\small

LENS & \textbf{Facts}: Penguins are native to deep, cold parts of southern hemisphere. Miami is located in the northern hemisphere and has a warm climate.
\textbf{Question:} Would it be common to find a penguin in Miami?
 \textbf{Rationale:} Where is a typical penguin's natural habitat? What conditions make \#1 suitable for penguins? Are all of \#2 present in Miami?
\textbf{Answer: No}
 \\\cline{2-2}
 & \textbf{Facts:} Shirley Bassey recorded the song Diamonds are Forever in 1971. Over time, diamonds degrade and turn into graphite. Graphite is the same chemical composition found in pencils.
\textbf{Question:} Is the title of Shirley Bassey's 1971 diamond song a true statement?
 \textbf{Rationale}: What is the title to Shirley Bassey's 1971 diamond song? Do diamonds last for the period in \#1? \textbf{Answer}: No \\\cline{2-2}
 & \textbf{Facts:} The first six numbers in Fibonacci sequence are 1,1,2,3,5,8. Since 1 is doubled, there are only five different single digit numbers.  \textbf{Question:} Are there five different single-digit Fibonacci numbers?
\textbf{Rationale:} What are the single-digit numbers in Fibonacci sequence? How many unique numbers are in \#1? Does \#2 equal 5?
\textbf{Answer:} Yes\\\cline{2-2}
  & \textbf{Facts:} Katy Perry's gospel album sold about 200 copies. Katy Perry's most recent pop albums sold over 800,000 copies.
\textbf{Question:} Do most fans follow Katy Perry for gospel music?
 \textbf{Rationale:} What type of music is Katy Perry known for? Is Gospel music the same as \#1?
 \textbf{Answer:} No\\\cline{2-2}
    & \textbf{Facts:} The Italian Renaissance was a period of history from the 13th century to 1600. A theocracy is a type of rule in which religious leaders have power. Friar Girolamo Savonarola was the ruler of Florence, after driving out the Medici family, from November 1494 â€ 23 May 1498. \textbf{Question:} Was Florence a Theocracy during Italian Renaissance?
  \textbf{Rationale:} When was the Italian Renaissance?When did Friar Girolamo Savonarola rule Florence? Is \#2 within the span of \#1? Did Friar Girolamo Savonarola belong to a religious order during \#3?
  \textbf{Answer:} Yes\\
  \midrule

\colorg \static{} & \colorg \textbf{Facts}: The average cost of a US Boeing 737 plane is 1.6 million dollars. Wonder Woman (2017 film) grossed over 800 million dollars at the box office. \textbf{Question}: Is a Boeing 737 cost covered by Wonder Woman (2017 film) box office receipts?
 \textbf{Rationale:}  How much does a Boeing 737 cost?. How much did the 2017 movie Wonder Woman gross? Is \#2 greater than \#1?
\textbf{Answer:} Yes
 \\\cline{2-2}\colorg
 & \colorg \textbf{Facts}: Big Show is a professional wrestler that weighs 383 pounds. Force is equal to mass times acceleration. An adult Cheetah weighs around 160 pounds. An adult Cheetah can run up to 58 MPH. \textbf{Question:}  Can a cheetah generate enough force to topple Big Show?
\colorg\textbf{Rationale}:  How much does Big Show weigh? How much does a cheetah weigh? How fast can a cheetah run? Is the force produced by a mass of \#2 and a speed of \#3 enough to knock over something that weighs \#1?
 \textbf{Answer}: Yes \\\cline{2-2}\colorg
 & \colorg \textbf{Facts}: Spaghetti and meatballs are a staple on Italian pizzeria menus in US. The Olive Garden, an Italian family restaurant, has several dishes with meatballs. Meatballs originated in the Chinese Qin dynasty (221 BC to 207 BC). \textbf{Question:} Do restaurants associate meatballs with the wrong country of origin?
 \textbf{Rationale:} In what country is the oldest evidence of people eating meatballs found? \dots Are \#3 and \#1 different?
\textbf{Answer:} Yes\\\cline{2-2}\colorg
&\colorg \textbf{Facts:} Torah scrolls must be duplicated precisely by a trained scribe. The Torah has a total of 8,674 words. The population of Bunkie Louisiana is 3,939 people according to a 2018 census. \textbf{Question:} Can you give at least one word from the Torah to all residents of Bunkie Louisiana?
\textbf{Rationale:} How many words are in the Torah? How many residents does Bunkie, Louisiana have? Is \#1 greater than \#2?
\textbf{Answer:} Yes\\\cline{2-2}\colorg
&\colorg \textbf{Facts:} Wrestlemania X took place in 1994. The Toyota Prius was first manufactured in 1997. \textbf{Question:} Could someone have arrived at Wrestlemania X in a Toyota Prius?
\textbf{Rationale:} When did Wrestlemania X hold? When was the Toyota Prius first manufactured? Is \#2 before \#1?
\textbf{Answer:} No\\
\midrule
\midrule
\end{tabular}
\caption{Qualitative analysis of exemplars for \textbf{\strat{}} dataset selected by LENS vs \static{}. Rationale is not completely shown for some questions to conserve space. However, in our experiments all exemplars include rationales.}
\label{tab:exemplar_qualitative_strategyqa}
\end{table*}